\documentclass[journal,]{IEEEtran}

\usepackage{amsmath,amsfonts}
\usepackage{amssymb}

\usepackage{algorithm}
\usepackage{algpseudocode}
\usepackage{array}
\usepackage[caption=false,font=footnotesize,labelfont=rm,textfont=rm]{subfig}
\usepackage{textcomp}
\usepackage{stfloats}
\usepackage{url}
\usepackage{verbatim}
\usepackage{graphicx}
\usepackage{cite}

\usepackage{amsmath}
\usepackage{tabularray}
\usepackage{multicol}
\usepackage{multirow}
\usepackage{xcolor}

	\newif\ifcomments\commentstrue

    \ifcomments    
    \newcommand{\ysc}[1]{{\ysa{*** #1 --YS}}}
    \else
    \newcommand{\ysc}[1]{}
    \endif
    
	\newif\ifcolors\colorstrue
	
	\ifcolors 
	\definecolor{darkred}{rgb}{0.7,0,0}
	\newcommand{\ysa}[1]{{\leavevmode\color{darkred}#1}} 
	\newcommand{\ysd}[1]{}  
 
	\else  
	
	\newcommand{\ysa}[1]{#1}
	\newcommand{\ysd}[1]{}   
	\fi

\begin{document}

\title{Visual-Semantic Graph Matching Net\\for Zero-Shot Learning}

\author{Bowen Duan, Shiming Chen, Yufei Guo, Guo-Sen Xie, Weiping Ding,~\IEEEmembership{Senior Member,~IEEE}\\ and Yisong Wang
\thanks{This work has been submitted to the IEEE for possible publication. Copyright may be transferred without notice, after which this version may no longer be accessible. (Corresponding author: Shiming Chen.)}
\thanks{Bowen, Duan and Yisong Wang are with the School of Computer Science and Technology, Guizhou University, Guizhou 550025, China and also with the Institute for Artificial Intelligence, Guizhou University, Guiyang, China (e-mail: gs.bwduan22@gzu.edu.cn; ys\_wang168@sina.com).} 
\thanks{Shiming Chen is with the Mohamed bin Zayed University of AI, Abu Dhabi 19282, UAE (e-mail: gchenshiming@gmail.com).} 
\thanks{Yufei Guo is with the Intelligent Science \& Technology Academy of CASIC, Beijing, China (e-mail: yfguo@pku.edu.cn).}
\thanks{Guo-Sen Xie is with the School of Computer Science and Engineering, Nanjing University of Science and Technology, Nanjing, China (e-mail: gsxiehm@gmail.com)}
\thanks{Weiping Ding is with the School of Artificial Intelligence and Computer Science, Nantong University, Nantong, 226019, China, and also the Faculty of Data Science, City University of Macau, Macau 999078, China (e-mail: dwp9988@163.com)}}



\maketitle

\begin{abstract}
Zero-shot learning (ZSL) aims to leverage additional semantic information to recognize unseen classes. To transfer knowledge from seen to unseen classes, most ZSL methods often learn a shared embedding space by simply aligning visual embeddings with semantic prototypes. However, methods trained under this paradigm often struggle to learn robust embedding space because they align the two modalities in an isolated manner among classes, which ignore the crucial class relationship during the alignment process. To address the aforementioned challenges, this paper proposes a Visual-Semantic Graph Matching Net, termed as VSGMN, which leverages semantic relationships among classes to aid in visual-semantic embedding. VSGMN employs a Graph Build Network (GBN) and a Graph Matching Network (GMN) to achieve two-stage visual-semantic alignment. Specifically, GBN first utilizes an embedding-based approach to build visual and semantic graphs in the semantic space and align the embedding with its prototype for first-stage alignment. Additionally, to supplement unseen class relations in these graphs, GBN also build the unseen class nodes based on semantic relationships. In the second stage, GMN continuously integrates neighbor and cross-graph information into the constructed graph nodes, and aligns the node relationships between the two graphs under the class relationship constraint. Extensive experiments on three benchmark datasets demonstrate that VSGMN achieves superior performance in both conventional and generalized ZSL scenarios. The implementation of our VSGMN and experimental results are available at github: \textcolor{blue}{https://github.com/dbwfd/VSGMN}

\end{abstract}

\begin{IEEEkeywords}

zero-shot learning, semantic-visual alignment, graph match, graph neural network

\end{IEEEkeywords}

\section{Introduction and Motivation}
\label{1}
\IEEEPARstart{S}{ince} the release of AlexNet~\cite{krizhevsky2012imagenet} in 2012, supervised deep neural networks have witnessed remarkable advancements~\cite{simonyan2014very,he2016deep,vaswani2017attention}. However, these highly effective architectures typically necessitate extensive, meticulously annotated datasets to achieve satisfactory performance. Furthermore, they often fail to recognize objects belonging to classes not included in the training set. In response to these limitations, zero-shot learning (ZSL) is introduced to mitigate these challenges~\cite{larochelle2008zero,palatucci2009zero,lampert2009learning}.

ZSL aims to train a model capable of transferring knowledge from seen classes to unseen ones by utilizing semantic information, thereby facilitating the classification of categories absent from the training set \cite{lampert2013attribute,fu2015transductive,fu2017zero}. Specifically, semantic information embeds both seen and unseen classes into high-dimensional prototypes, which can be attribute vectors manually defined \cite{lampert2009learning} or automatically extracted word vectors \cite{socher2013zero}, among other representations. In essence, ZSL leverages semantic information to bridge the gap between seen and unseen classes. According to their classification range, ZSL methods can be categorized into conventional ZSL (CZSL), which aims to predict unseen classes, and generalized ZSL (GZSL), which can predict both seen and unseen classes \cite{xian2017zero}.

\begin{figure}[!t]
    \centering
    \subfloat[ Existing visual-semantic embedding method]{ \includegraphics[width=3.5in]{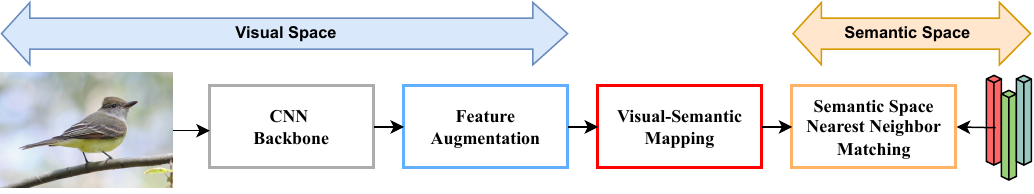} \label{fig:1a}}
    \quad
    \subfloat[ Existing category-relation based method]{ \includegraphics[width=3.5in]{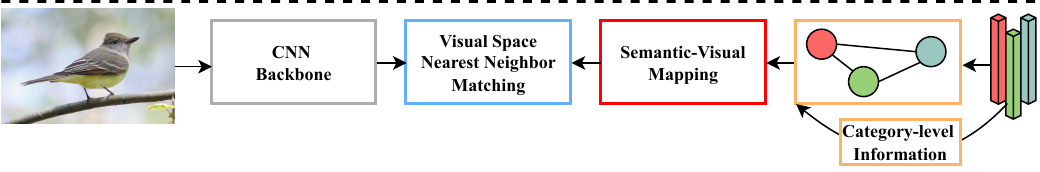} \label{fig:1b}}
    \quad
    \subfloat[Our VSGMN]{ \includegraphics[width=3.5in]{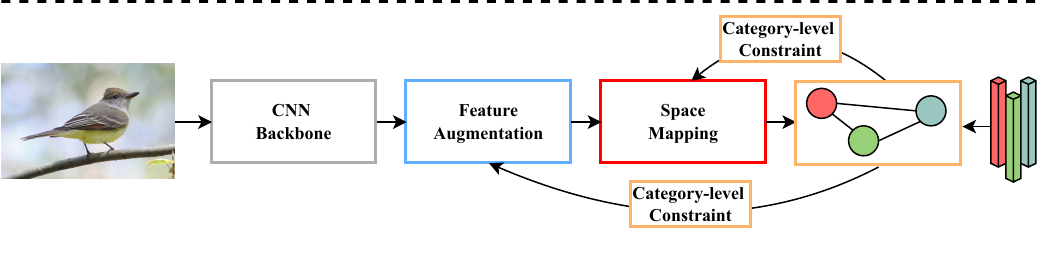} \label{fig:1c}}
    \caption{Motivate illustration. (a) Most existing embedding-based methods treat semantic vectors solely as classifiers, neglecting the crucial inter-class information ($e.g.$, the relationship between a cat and a lion is much closer than the relationship between a cat and a bird) inherent in semantic vectors. (b) Existing methods based on category relationships, although attempting to explore category relationship information in semantic vectors, often confine this utilization to the semantic space. (c) Our VSGMN not only utilizes the category relationship information in semantic vectors but also transfers this information from the semantic space to the visual space. This enables us to impose class-level relationship constraints on the augmentation of visual features and space mappings.}
    \label{fig:1}
\end{figure}

    \label{fig:1}

Currently, in the field of ZSL, there are two groups of approaches: generative methods and embedding methods.
Generative methods \cite{schonfeld2019generalized,chen2021hsva,xian2019f,yu2020episode,chen2021free, chen2023egans,hong2022semantic,chen2023evolving} employ generative models to generate unseen visual features and convert ZSL into traditional supervised learning. By contrast, embedding-based ZSL methods\cite{hou2024visual,chen2022msdn,chen2022transzero} aims to train a model to learn an embedding space where visual features are associated with their corresponding semantic prototypes. These methods identify unseen classes by calculating the similarity between feature embeddings and different class prototypes in the learned space.\\

However, most ZSL models align visual and semantic features merely by minimizing the distance between embeddings and their corresponding prototypes, which ignores the alignment of category relationships in the semantic space. This leads to an inconsistency between the distance relationships among prototypes and embeddings in the semantic space. Even though the embeddings are close to their prototypes, the mismatch in relationships makes the learned embedding space more prone to class confusion. In these methods, most embedding-based ZSL approaches \cite{xie2020region,chen2022transzero} are prone to this relationship mismatch issue because, in these approaches, semantic prototypes often serve merely as more intricate labels to guide the direction of the model optimization process, as illustrated in Fig. \ref{fig:1} (a). Some alternative methods \cite{wang2018zero,li2019rethinking} utilize the relationships between semantic prototypes to learn semantic-visual mapping. Nevertheless, in these methods, the information regarding category relationships is confined to the semantic space, which leads to the unresolved issue of mismatched visual-semantic relationships, as depicted in Fig. \ref{fig:1} (b). These limitations pose challenges for ZSL methods to acquire a precise visual-semantic mapping that adequately addresses both seen and unseen classes. 

To address the aforementioned challenges, in this paper, we propose the Visual-Semantic Graph Matching Net, termed as VSGMN. As depicted in Fig. \ref{fig:1} (c), VSGMN leverages the relationships within the semantic space to constrain the augmentation of visual features and space matching, thereby guiding the model to correctly match relationships and ultimately learn a robust embedding space. Specifically, VSGMN consists of two main components: the graph build net (GBN) and the graph matching net (GMN). In the GBN, VSGMN generates the virtual unseen features based on the semantic relation and utilizes a visual-semantic embedding network to obtain semantic embeddings and achieve the first-stage visual-semantic alignment by making the embedding closer to its prototype. These embeddings and its corresponding prototypes are used to build initial visual and semantic graphs in the semantic space for subsequent processing by the GMN, respectively. Our GMN, on the other hand, comprises two interconnected branches: the visual branch and the semantic branch. Each branch initializes its data based on the graph built in GBN, and employs graph neural networks (GNNs) to encode structural information between nodes and the differential information between the two graphs into the node representations. Finally, guided by the class relationship constraint loss, we constrain the visual-semantic embedding network in GBN, to achieve second-stage alignment of vision and semantics.

The main contributions of this paper are summarized as:
\begin{itemize}[]
\item We propose a novel ZSL method called VSGMN to tackle the issue of missing semantic category relationship information in visual-semantic embedding. VSGMN consists of a GBN and a GMN that achieve different stages alignment between visual and semantic respectively, thus further boosting the performance of ZSL.
\item We propose a graph match layer to alleviate the visual-semantic gap caused by inconsistent manifolds and matching different-order relationships between two built graphs, which achieve second-stage visual-semantic alignment. This graph match layer is incorporated into GMN. 
\item Extensive experiments on three challenging benchmark datasets, $i.e.$, AWA2 \cite{xian2017zero}, CUB \cite{welinder2010caltech} and SUN \cite{patterson2012sun}, demonstrate the superior performance of VSGMN. Compared with its baseline \cite{chen2022transzero}, VSGMN leads to significant improvements of 2.2\%/0.9\%, 2.4\%/1.3\% and 1.4\%/0.3\% in acc/H on the three benchmarks, respectively.
\end{itemize}

The rest of this paper is organized as follows. Sec. \ref{2} discusses
related works. The proposed VSGMN is illustrated in Sec. \ref{3}.
Experimental results and discussions are provided in Sec. \ref{4}. Finally,
we present a summary in Sec. \ref{5}.
\section{Related Work}
\label{2}
\subsection{Relation based Zero-Shot Learning}
Early ZSL methods\cite{song2018transductive,li2018discriminative,li2019rethinking,xian2018feature,yu2020episode,min2020domain} aim to establish a robust space mapping connecting visual and semantic spaces, enabling the flow of semantic information from seen to unseen categories. These methods typically employ pre-trained or end-to-end trainable networks to extract visual features and utilize loss functions such as cross-entropy to bring the visual embeddings closer to their semantic prototypes, thereby achieving visual-semantic alignment. Despite their ability to recognize unseen classes, these methods often yield relatively sub-optimal results due to overlooking the inherent category information in ZSL tasks. A similar approach to this work is relation-based zero-shot methods, which can typically be categorized into attribute relationship-based zero-shot methods and category relationship-based zero-shot methods. Methods based on attribute relationships typically employ attention mechanisms to localize attribute regions \cite{xie2019attentive,huynh2020fine,xie2020region,chen2022transzero,chen24gndan,chen2022transzero++,ge2022dual,chen2024rethinking,chen2024progressive,chen2024causal} or use GNNs for attribute reasoning \cite{xie2020region,chen24gndan,guo2023graph}. Regardless of the approach, the focus of these methods typically lies in augment visual features by leveraging relationships. Another type of methods based on category relationships \cite{wang2018zero,li2019rethinking} often emphasize the semantic space, typically utilizing GNNs to generalize semantic representations of unseen classes in the visual or latent space.

Unfortunately, these methods, while attempting to leverage relationships in the model, are limited either to the visual or semantic domain. Due to the inherent nature of ZSL tasks, models often need to establish connections between two different modalities of feature spaces. Isolated utilization of relationships can be effective, but for the crucial visual-semantic connections, it's often necessary to jointly utilize both spaces to achieve better results. For example, utilizing category relationships in the semantic space to constrain visual feature enhancement and mapping. One major constraint hindering this approach stems from the fact that the class membership of samples within the model is unknown, making it difficult to impose class constraints on the visual space.

In contrast, we propose a novel network architecture that utilizes category relationships in the semantic space to guide the learning of visual space, enabling the model to learn visual representations and visual-semantic embeddings that better conform to the manifold structure of the semantic space.
\subsection{Visual-Semantic Gap}
ZSL leverages semantic vectors to classify unseen visual samples. In this process, there exist not only two distinct domains of unseen and seen classes but also a significant gap between the semantic and visual domains. Visual and semantic modalities are embedded on different manifold structures, making it highly challenging to utilize information from the semantic space to classify samples in the visual space, including unseen class samples. Therefore, the critical task of ZSL is to learn a visual-semantic alignment.

To address this issue, some ZSL methods \cite{frome2013devise,wang2017zero,hubert2017learning,liu2018generalized,schonfeld2019generalized,chen2021hsva} attempt to construct a common latent space to bridge the gap. Other methods \cite{xian2017zero,li2019rethinking}, however, argue that compared to the semantic space, the visual space often possesses more valuable discriminative power. Consequently, they construct mappings from the semantic space to the visual space. However, methods that involve constructing a latent space often require more complex constraints and network structures to achieve satisfactory results due to the mapping between three spaces. On the other hand, methods focused on constructing semantic-visual mappings only aim to preserve more information in the embedding vectors from the perspective of information loss and do not directly reduce the gap between the two spaces. Therefore, these models still suffer from the projection domain shift and bias problems \cite{pourpanah2022review}.

In contrast, we choose to establish the most fundamental visual-semantic mapping. However, we strive to leverage the inherent relationships among semantic vectors to constrain the augmentation of visual features and space mapping from the visual space. This aims to achieve the second-stage alignment between the visual and semantic spaces, thereby mitigating the space gap.

\subsection{GNNs for Graph Matching}
The history of GNNs goes back to at least the early work \cite{gori2005new} in 2005, which propose to use a propagation process to learn node representations. With the rise in popularity of deep convolutional networks, many researchers aspire to apply the advantages of convolutional networks to GNNs, known as Graph convolutional Networks (GCN). Some methods focus on constructing spatial-domain graph convolution operations \cite{hamilton2017inductive}, referred to as spatial GCN. Others \cite{bruna2013spectral,defferrard2016convolutional} utilize Fourier transforms to extract patterns in the spectral domain, known as spectral GCN. Recently, due to the popularity of attention mechanisms, there has been research \cite{velickovic2017graph} integrating attention mechanisms into GNNs, known as graph attention networks (GAT). Due to the nature of the data types, these GNNs are inherently required to have the ability to extract structural features. Consequently, GNNs have been widely applied in various fields such as node classification, edge prediction, graph embedding, and graph matching.

Specifically, the graph matching problem requires establishing correspondences between nodes in two or more graphs. This field often leverages the advantages of GNNs to handle graph-structured data and then perform matching between nodes. Motivated by this, we consider aligning the visual space and semantic space in ZSL as a graph matching task between a visual graph and a semantic graph. Specifically, we use GNNs to update the node representations in both the visual graph and the semantic graph, incorporating higher-order neighborhood nodes information and cross graph information. Next, we align the visual graph with the semantic graph to constrain the visual-semantic embedding, thereby achieving better alignment between the visual space and the semantic space.

\begin{figure*}[!t]
    \centering
    \includegraphics[width=\textwidth]{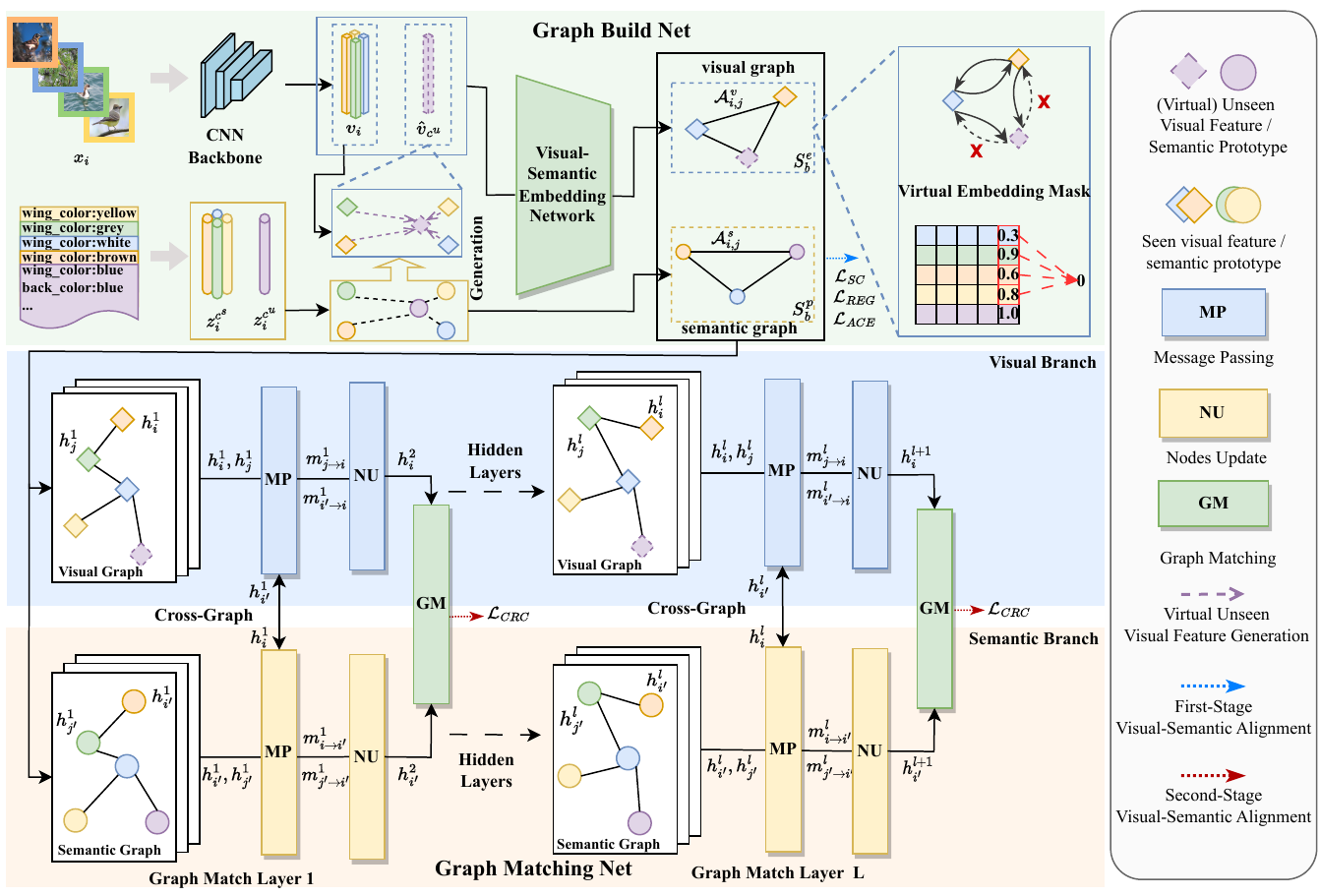}
    \caption{The architecture of the proposed VSGMN model. VSGMN consists of a GBN and a GMN. GBN aims to build the visual graph and semantic graph in the semantic space. To ensure the dimensions of node representations are the same for both, and to achieve the first-stage alignment between vision and semantics, we use the visual-semantic embedding network ($e.g.$, TransZero \cite{chen2022transzero}) to bridge the two spaces. Additionally, we propose a method to build virtual features for unseen classes in this stage to utilize the relationships among unseen classes and mask the generated virtual embeddings to prevent interference from noise. GMN constrains the training processes of GBN by matching the category relationship between visual embeddings and semantic prototypes.  }
    \label{fig:2}
\end{figure*}

\section{Proposed Method}
\label{3}
In this section, we describe our model VSGMN for ZSL in detail, starting with some notations and the problem definition.

\subsection{Problem Description}
Assume that we have training data $D^s=\{(x_i^s,y_i^s,z_i^s)_{i=1}^{N_s}\vert x_i^s\in X^s,y_i^s\in Y^s,z_i^s\in Z^s \}$ with $C_s$ seen classes and test data $D^u=\{(x_i^u,y_i^u,z_i^u)_{i=1}^{N_u}\vert x_i^u\in X^u,y_i^u\in Y^u,z_i^u\in Z^u \}$ with $C_u$ unseen classes where $x_i$ denotes the image i, $y_i$ is the corresponding class label and $z_i$ is the semantic class prototype which helps information transfer from seen to unseen classes. The semantic class prototypes $z_i \in \mathbb{R}^K$ are either manually defined attribute vectors or learned from language models. Note that $Y^s \cap Y^u = \varnothing$ and $X = X^s \cup X^u$. We also use the semantic attribute word embeddings $a_i$ of each attribute learned by GloVe\cite{pennington2014glove} . The goal of CZSL is to learn a classifier for unseen images, $i.e.$, $f_{CZSL} : X \to Y^u$, while for GZSL the goal is to learn a classifier for seen and unseen images, $i.e.$, $f_{GZSL} : X \to Y^u \cup Y^s $.

\subsection{Overview}
In this paper, we propose the VSGMN to leverage category-level relationships in semantic prototypes to aid in semantic-visual alignment, thereby enhancing the quality of the embeddings from visual to semantic space, which alleviate the visual-semantic gap problem. As illustrated in Fig.\ref{fig:2}, our VSGMN comprises a GBN, and a GMN. In the GBN, we first build virtual visual features for unseen classes and combine them with the training samples of seen classes. Through the visual-semantic embedding network, we obtain semantic embeddings of visual features, and achieve first-stage visual-semantic alignment by aligning embedding with its prototypes. Next, we build visual and semantic graphs in the semantic space. To ensure the credibility of the visual graph, we apply a virtual embedding mask to it. The whole GBN is constrained by GMN, a dual-branch network intended to achieve the second-stage alignment between visual and semantic by matching the two graphs. 
\subsection{Graph Build Net}
\label{3.1}
In GBN, our task is to provide reliable node representations and structural information for the initial visual and semantic graphs to be used in the subsequent graph matching process for relationship constraints. Since the dimensions of the visual space and semantic space are not the same, directly initializing the graph structures in their respective spaces would make it difficult to utilize relationship information in the subsequent stages. Therefore, we choose to utilize existing embedding methods to obtain semantic embeddings, achieving dimension unification and the first-stage visual-semantic alignment by aligning the embeddings with its semantic prototypes. In addition, to provide richer category relationship information, we also build virtual unseen class visual features at this stage to leverage unseen category relationships. Meanwhile, to maintain the credibility of node features and structural features, virtual unseen embedding mask is introduced during the construction of the visual graph to eliminate noise.

\subsubsection{Virtual Unseen Visual Feature Generation}
Due to the nature of ZSL tasks, the model must establish connections not only between the visual and semantic spaces but also between seen and unseen classes. This implies that the category relationships we aim to leverage should encompass not only seen classes but also include relationships between seen-unseen classes and unseen-unseen classes. Since during the training phase, the model can simultaneously utilize semantic prototypes of both seen and unseen classes, which means that we can easily introduce information about unseen classes into the semantic graph. Thus, it constrains us from leveraging unseen class relationships solely in the visual domain. Based on these considerations, we propose generating virtual unseen class visual features to leverage the unseen class relationships inherent in semantic prototypes.

Specifically, we adopt a method similar to \cite{santoro2016meta,snell2017prototypical,liu2021isometric}. First, we build visual feature prototypes $P_{c}$ for each seen class $c_s$ by calculating the mean of the samples belonging to class $c$:
\begin{equation}
P_{c}=\frac{1}{n_{c}}\sum_{i=1}^{n_{c}}v_i^{c},  
\end{equation}
where $v_i^{c} \in \mathbb{R}^{HW\times C}$  is the visual feature extracted by the CNN backbone \cite{he2016deep} for the $i$-th image in class $c_s$ and $n_{c}$ is the total number of samples in class $c_s$.

After obtaining the visual prototypes for the \( C_s \) seen classes, we build virtual visual features \( \hat{v}_{c^u} \) for each unseen class \( c_u \):
\begin{equation}
\hat{v}_{c^u}=\frac{1}{k}\sum_{i \in \mathcal{N}_{c}}{P_{i}}, 
\label{eq: 2}
\end{equation}
where $\mathcal{N}_{c}$ is the set of the top-$K$ ($K$ is a hyperparameter) seen classes with the largest cosine similarities to class ${c_u}$ in semantic prototype space $Z$.

During the subsequent training phase, the virtual unseen class visual features and seen training samples will be projected to semantic space by the visual-semantic embedding network in GBN, which achieve the first-stage visual-semantic alignment by aligning the seen class semantic embedding with its semantic prototypes. Specifically, each batch of samples contains both samples extracted from the seen class training set and all the virtual unseen class visual feature generated in this phase:
\begin{equation}
    v_{b}=\{v_1,v_2,\dots,v_{n_b},\hat{v}_{n_b+1},\hat{v}_{n_b+2},\dots,\hat{v}_{n_b+C_u}\}.
\end{equation}
\subsubsection{Graph Build with Virtual Unseen Embedding Mask}
After the previous section, we obtain semantic embeddings with dimensions identical to those of the semantic prototypes, which serve as our visual graph node representations (prototypes as semantic node representations). Next, we provide the structure information (edge weights) for visual graph and semantic graph by compute the weighted adjacency matrix $\mathcal{A}^v$ and $\mathcal{A}^s$.
It is defined as:   
\begin{align}
    \label{eq:4}&\mathcal{A}^v = {\Bar{S}}^{e^{\top}}_{b}\Bar{S}_b^e, \\
    \label{eq:5}&\mathcal{A}^s = {\Bar{S}}^{p^{\top}}_{b}\Bar{S}_b^p,\\
    &\Bar{S}_b^e=[\Bar{s}_1,\Bar{s}_2,\dots,\Bar{s}_{n_b},\Bar{\hat{s}}_{n_b+1},\Bar{\hat{s}}_{n_b+2},\dots,\Bar{\hat{s}}_{n_b+C_u}],\\
    &\Bar{S}_b^p=[\Bar{z}_1,\Bar{z}_2,\dots,\Bar{z}_{n_b},\Bar{z}_{n_b+1},\Bar{z}_{n_b+2},\dots,\Bar{z}_{n_b+C_u}],
\end{align}
where $\Bar{s}_i$, $\Bar{z}_i \in \mathbb{R}^{D}$  is the normalized semantic embeddings from $v_i$ and its corresponding standardized semantic prototype. Each element in $\mathcal{A}^v$ or $\mathcal{A}^s$ represents the cosine distance between the respective two semantic embeddings or prototypes.

Note that both the visual graph and the semantic graph built by us contain nodes corresponding to all unseen classes. In the semantic graph, the nodes representing unseen classes are authentic and reliable, as they originate from our prototypes of unseen classes. However, for the visual graph, these unseen visual features are derived from seen visual features based on semantic relationships, which often leads to discrepancies from true unseen features. Therefore, without modification, directly utilizing them to build the visual graph would introduce noise to the propagation process, impacting the reliability of our seen class embeddings and subsequently affecting the relationship matching process. 

Inspired by \cite{zhang2023attribute}, we propose a virtual embedding mask to eliminate noise from virtual classes, preserving real classes while still incorporating valuable unseen category information in the subsequent graph matching process. Specifically, for visual graph, we remove the out-degree of unseen embeddings while preserving their in-degree:
\begin{equation}
\label{eq:8}
    \mathcal{A}^m_{i,j} = 
    \begin{cases}
        0 & \mathrm{if}\ j \in [n_b+1,n_b+C_u],\\ 
        {A}^v_{i,j} & \mathrm{otherwise}.
    \end{cases}  
\end{equation}
Then, we obtain the masked adjacency matrix $\mathcal{A}^m$, which represents the neighborhood relationships in the visual graph and is subsequently used in the visual branch.
\begin{figure*}[!t]
    \centering
    \includegraphics[width=\textwidth]{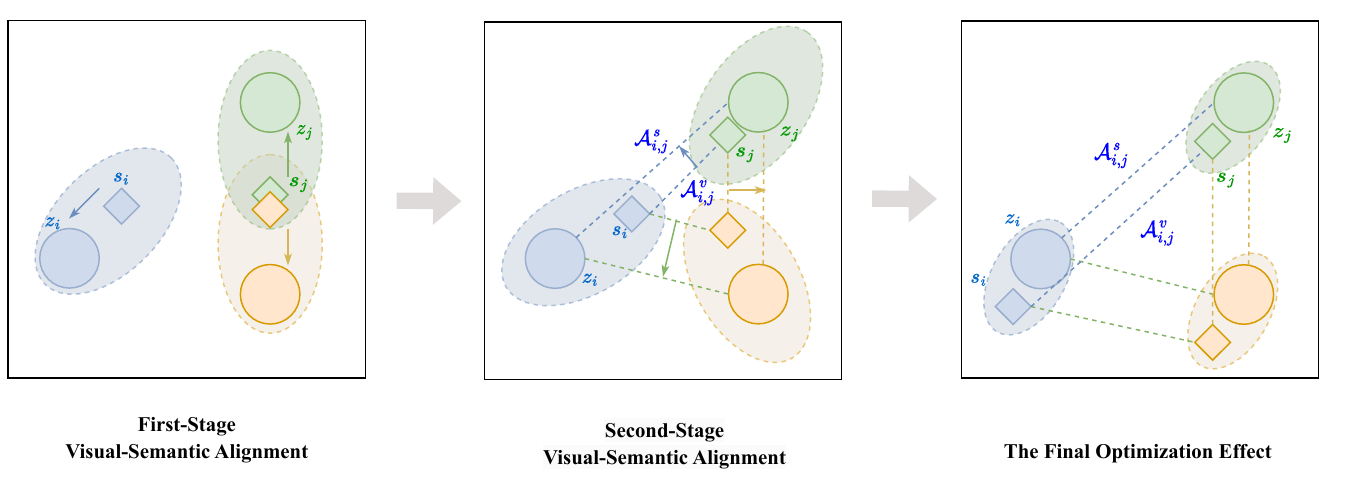}
    \caption{The first-stage visual-semantic alignment, the second-stage visual-semantic alignment and the optimization effect achieved by simultaneously using both constraint methods. The shaded area represents the region where samples of this category may appear, while the dashed line indicates the inter-class relationships that need to be aligned with each other between the two graphs. }
    \label{fig:3}
\end{figure*}
\subsection{Graph Matching Net}
Typically, after obtaining semantic embeddings of visual feature, most ZSL methods match these semantic embeddings with their corresponding semantic prototypes through loss function such as cross-entropy. We utilize the same strategy in GBN. Essentially, the semantic embeddings can be seen as predictions of class labels for the images in these embedding-based ZSL approach, where each semantic prototype can be seen as a more complex class label than a one-hot label. Therefore, after obtaining the semantic embeddings, the majority of the ZSL methods are essentially concluded. However, the effectiveness of most ZSL models trained according to this paradigm is often unstable. The reason lies in the visual-semantic gap mentioned earlier, the gap between the visual space and the semantic space make it difficult to bridge through a single matching process.

To alleviate the aforementioned issues, we propose a second-stage visual-semantic alignment through graph matching approach. By aligning the class relationships among the semantic embeddings of visual features $S_b^e$  with the class relationships among the semantic prototypes $S_b^p$, we constrain the learning process of space embedding after the first-stage visual-semantic alignment in GBN. This ensures that the overall network model learns more appropriate space mapping function to reduce the visual-semantic gap as shown in Fig.\ref{fig:3}.

Specifically, we propose a graph matching network based on GNNs to achieve the aforementioned process. This network consists of two interconnected branches: the visual branch and the semantic branch. Each branch is composed of multiple graph matching layers stacked together. In each layer of our graph matching network, we incorporate the structural information (relationship information) inherent in the node graph, the semantic information carried by the node features themselves, and the cross-graph discrepancy information through propagation, into the node representations. Finally, by aligning the relationships between visual nodes and semantic nodes at each layer, we achieve the second-stage alignment between visual and semantic spaces.

\subsubsection{Graph Match Layer}
\label{3.4.1}
To achieve better matching between vision and semantics, we intend to perform the second-stage visual-semantic alignment through aligning category relationships between the two built graphs. However, directly aligning the original semantic embedding graph with the semantic prototype graph often does not yield good results. We believe that this phenomenon is mainly due to two reasons. i) Although visual features are embedded into the semantic space, they still reside in different manifolds. Therefore, directly aligning category relationships using distance metrics may not yield good results. ii) Directly aligning relationships can only achieve first-order relationship matching but it cannot handle higher-order relationship matching (relationships between nodes and their higher-order neighbors).

Therefore, we propose a graph match layer to perform intra-graph and inter-graph information propagation between the visual graph and the semantic graph. Intra-graph information propagation leverages the relationship information between nodes to automatically integrate information from both first-order and higher-order neighbors into node representations. On the other hand, inter-graph information propagation helps alleviate biases caused by inconsistent manifold structures.

Specifically, each graph match layer consists of three processes: message passing, nodes updating, and graph matching. Since the visual branch and the semantic branch share the same processing steps except for the initial graph data input, in this section, we only introduce the visual branch. The semantic branch can be analogously derived from this.

\noindent \textbf{Message Passing.} During this process, the graph matching layer computes the message each node should receive in this layer, which includes message from neighboring nodes within the graph as well as cross-graph differential information from the other graph. Specifically, it utilizes the message passing net  $f_{\mathrm{message}}^{\mathrm{i}}$, $f_{\mathrm{message}}^{\mathrm{c}}$ to compute the information that nodes need to receive. It is defined as:
\begin{align}
    \label{eq:9} &m_{j\rightarrow i} = f_{\mathrm{message}}^{\mathrm{i}}(h_i^l,h_j^l,\mathcal{A}^m_{i,j}),\ j \in \mathcal{N}_i,\\
    \label{eq:10} &m_{{i'}\rightarrow i} = f_{\mathrm{message}}^{\mathrm{c}}(h_i^l,h_{i'}^l),
\end{align}
where $h_i^l$, $h_{i'}^l \in \mathbb{R}^{D}$ is the $i$-th node  representation of the $l$-th visual graph match layer and semantic graph match layer respectively. Specifically, when $l=1$, $h_i^l = s_i$, $h_{i'}^l = z_i$. $\mathcal{N}_i$ is the neighbours defined in $\mathcal{A}^m$, $m_{j\rightarrow i}$ is the intra-message passed from the $j$-th node to the $i$-th node of the visual graph and $m_{{i'}\rightarrow i}$ is the cross-message passed from semantic graph to visual graph.

\noindent \textbf{Nodes Updating.} In this process, the graph match layer utilizes the information collected during the previous process to update each node in the visual graph, incorporating neighborhood information and cross-graph discrepancy information into the representation of each node. Specifically, it uses the node update functions $f_{\mathrm{node}}$ to aggregate these information and obtain new node representations. It is defined as:
\begin{align}
\label{eq:11}
    &h_i^{l+1}=f_{\mathrm{node}}(h_i^l, M_{j\rightarrow i},m_{{i'}\rightarrow i}),
\end{align}
where $M_{j\rightarrow i}$ is the pack of intra-graph neighborhood message of the $i$-th node. $m_{{i'}\rightarrow i}$ is the cross-graph message.

Notably, in the graph matching layer, after the processes of message passing and nodes updating, each node naturally acquires the necessary information from its first-order neighbors within the current layer's graph. Consequently, as the number of layers increases, the degree of neighbor information incorporated into each node also increases ($e.g.$, in the second layer, the visual and semantic graphs effectively integrate second-order neighbor information from the initial graph). This progression lays the groundwork for the subsequent graph matching stage, enabling the alignment of multi-order neighbor relationships.

\noindent \textbf{Graph Matching.} In the first two stages, graph match layers are guided by the structural information of the graph to integrate semantic information from neighbors and cross-graph discrepancy information into each node, resulting in a new visual graph and semantic graph. Each node automatically incorporates richer information, making the relationships extracted from these two graphs more reliable. Therefore, the graph match layer chooses to use the updated graphs as the basis for graph matching. Specifically, in the graph matching process, it first computes the relationships between nodes within each graph, and then utilizes class constraint losses to align the relationships between the two graphs, thereby achieving the second-stage alignment. It is defined as:
\begin{align}
    &p^{sl}_{i,j}=\frac{\mathrm{exp}(\Bar{h}_i^{s(l+1)} \times \Bar{h}_j^{s(l+1)})}{\sum_{\hat{j}=1}^{n_b+C_u}\mathrm{exp}(\Bar{h}_i^{s(l+1)} \times \Bar{h}_{\hat{j}}^{s(l+1)})},\\
    &p^{vl}_{i,j}=\frac{\mathrm{exp}(\Bar{h}_i^{v(l+1)} \times \Bar{h}_j^{v(l+1)})}{\sum_{\hat{j}=1}^{n_b+C_u}\mathrm{exp}(\Bar{h}_i^{v(l+1)} \times \Bar{h}_{\hat{j}}^{v(l+1)})},
\end{align}
where $p^{sl}_{i,j}$ and $p^{vl}_{i,j}$ represent the probability distribution of the relation score between updated nodes $h_i^{l+1}$ and $h_j^{l+1}$ in the $l$-th graph match layer of the semantic graph and visual graph, respectively. $\Bar{h}_i^{v(l+1)}$, $\Bar{h}_i^{s(l+1)}$ is the node representation after being L2 normalized by $h_i^{l+1}$ from updated visual and semantic graph respectively. Note that, we only present the relationship extraction part here. The specific class constraint loss functions will be demonstrated in subsequent section.
\subsubsection{Specific Implementation For Graph Match Layer}
\label{3.2.2}
For the message passing function $f_{\mathrm{message}}$ and node update function $f_{\mathrm{node}}$, we propose two different implementations. One is based on attention, while the other is based on propagation similar to \cite{li2019graph}.

\noindent {\bf{Attention-Based Implementations.}} Due to the graph attention mechanism's ability to automatically learn the importance of different neighbors, we choose a mechanism similar to the one used in \cite{velickovic2017graph,liu2021isometric} as our graph match layer. Specifically, for the implementation based on attention mechanism, we first map the node representations to a latent space. Then, we compute the similarity of node representations in the latent space, normalize it to obtain attention scores, and finally update the nodes by weighted summation based on the attention scores. This process is defined as:

\begin{align}
     &\beta_{j\rightarrow i}^l=\mathrm{cosdis}(W_m^l  h_i^l,W_m^l  h_j^l),\\
     &\alpha_{j\rightarrow i}^l=\mathrm{softmax}(\tau\beta_{j\rightarrow i}^l),\\
    &m_{j \rightarrow i} = ={\alpha}_{j \rightarrow i}^l h_j^l,\\
    &m_{{i'}\rightarrow i} = h_i^l - h_{i'}^l,\\
    &h_i^{l+1}=\sum\limits_{j \in \mathcal{N}_i}m_{j\rightarrow i} + \lambda m_{{i'}\rightarrow i},
\end{align}
where $\beta_{j\rightarrow i}^l$ and $\alpha_{j\rightarrow i}^l$ is the similarity score and attention score, respectively, between node $j$ and node $i$ of the $l$-th layer. $W_m^l$ is the learnable matrices of weights, $\mathrm{cosdis}(\cdot,\cdot)$ is a function to compute the cosine distance, $\tau$ and $\lambda$ are two hyperparameters.

\noindent {\bf{Propagation-Based Implementations.}} Compared to the attention-based implementation, the propagation-based implementation can preserve more cross-graph information differences. However, because we have only excluded the out degree of unseen categories in the visual graph without imposing additional constraints on neighbor selection, the propagation-based version may be weaker in the process of intra-graph information pass. Specifically, we use MLP to get the passed message. Afterward, we concatenate the the node itself, intra-information and cross-difference and  pass them into the node updating network to ultimately obtain the new visual features. It is defined as:

\begin{align}
    &m_{j \rightarrow i} = ReLu\Bigl(W_{pm}^l\Vert(h_i^l,h_i^l,e_{ij})+b_{pm}^l\Bigr), \\
    &m_{{i'}\rightarrow i} = h_i^l - h_{i'}^l,\\
    &h_i^{l+1}=ReLu\Biggl(W_{pn}^l\Vert\Bigl(h_i^l,\sum\limits_{j \in \mathcal{N}}m_{j \rightarrow i},m_{{i'}\rightarrow i}\Bigr)+b_{pn}^l\Biggr),
\end{align}
where $W_{pm}^l$,$b_{pm}^l$,$W_{pn}^l$,$b_{pn}^l$ are the weights and biases of the linear layers respectively, $e_{ij}$ is a vector flattened by $\mathcal{A}^m_{i,j}$ which represents the edge features, and $\Vert$ is the concatenate function. Besides, there is a LayerNorm layer behind the message passing net and nodes Update net.

Note that, regardless of the specific implementations, each graph match layer will facilitate the flow of information within both graphs, and the range of information aggregated by graphs generated in different layers will be different. Generally, in deeper layers, the feature representation of each node will contain the information from higher-order neighbors. Therefore, in the subsequent graph alignment process, we will match each layer of the graph separately, thereby applying different-order relationship constraints to visual-semantic embedding simultaneously to enable them to learn more robust inter-class relationships.
\subsection{Model Optimization}
To achieve better alignment of vision and semantics using class-level relationships, our proposed VSGMN incorporates not only the common first-stage visual-semantic alignment loss functions used in ZSL methods—regression loss function or cross-entropy loss function but also an additional second-stage visual-semantic alignment loss function based on KL divergence: the class relationship constraint loss. 

\subsubsection{First-Stage Visual-Semantic Alignment}
\label{3.5.1}
In First-Stage Visual-Semantic Alignment, the loss functions will make the semantic embedding $s_i$ close to its semantic prototype $z_i$. Specifically, we use MSE loss and cross-entropy loss to achieve the first visual-semantic matching.

\noindent\textbf{MSE-Based Regression Loss.} To encourage GBN to accurately map the visual features into their corresponding semantic prototypes, we introduce the MSE based regression loss to constrain VSGMN:
    \begin{equation}
        \mathcal{L}_{REG}=-\frac{1}{n_b}\sum_{i=1}^{n_b}\Vert s_i-z_i \Vert_2^2.
    \end{equation}

\noindent\textbf{Attribute-Based Cross-Entropy Loss.} To enhance the distinctiveness of visual features between different classes and further match the semantic embeddings with their corresponding semantic prototypes, we employed a Attribute-Based cross-entropy loss to further achieve first-stage visual-semantic alignment:
    \begin{equation}
        \mathcal{L}_{ACE}=-\frac{1}{n_b}\sum_{i=1}^{n_b}\log{\frac{\mathrm{exp}(s_i \times z_i)}{\sum_{c \in C}\mathrm{exp}(s_i \times z^c)}}.
    \end{equation}
    
\noindent\textbf{Self-Calibration Loss.} Due to the constraints introduced by $\mathcal{L}_{REG}$ and $\mathcal{L}_{ACE}$, which only exist within the seen classes, the first-stage visual-semantic constraints we apply are inevitably overfitted to the seen classes, as also observed in \cite{zhu2019semantic,huynh2020fine,xu2020attribute,chen2022transzero}, thereby affecting the subsequent second-stage constraints. To address this and introduce unseen class information into the first-stage constraints, we utilized Self-Calibration Loss proposed by\cite{chen2022transzero}, which explicitly shifts some of the prediction probabilities from seen to unseen classes. It is defined as:
    \begin{equation}
        \mathcal{L}_{SC}=-\frac{1}{n_b}\sum_{i=1}^{n_b}{\sum_{c' \in C^u} }\log{\frac{\mathrm{exp}(s_i) \times z^{c'} + \mathbb{I}_{[c' \in C^u]})}{\sum_{c \in C}\mathrm{exp}(s_i) \times z^c+ \mathbb{I}_{[c' \in C^u]})}},
    \end{equation}
where $\mathbb{I}_{[c \in C^u]}$ is an indicator function ($i.e.$, it is 1 when $c \in C^u$, otherwise -1).
\begin{algorithm}[!t]
  \caption{ The algorithm of VSGMN.}
  \label{alg:1}
  \label{alg:Framwork}
  \begin{algorithmic}[1]
    \Require
      The training set $D^s=\{(x_i^s,y_i^s,z_i^s)_{i=1}^{N_s}\vert x_i^s\in X^s,y_i^s\in Y^s,z_i^s\in Z^s \}$,
      the test set $D^u=\{(x_i^u,y_i^u,z_i^u)_{i=1}^{N_u}\vert x_i^u\in X^u,y_i^u\in Y^u,z_i^u\in Z^u \}$,
      the pretrained CNN backbone ResNet101, the maximum iteration epoch $\mathrm{max}_{iter}$, loss weights ($i.e.$, $\mathcal{L}_{REG}$, $\mathcal{L}_{SC}$, $\mathcal{L}_{CRC}$, $\lambda_{REG}$, $\lambda_{SC}$, $\lambda_{CRC}$), and hyperparameters (learning rate = 0.001, momentum = 0.9, weight decay = 0.0001) of the SGD optimizer.
    \Ensure
      The predicted label $c^*$ for the test samples.
    \State $iter \leftarrow 0$;
    \While {$iter < \mathrm{max}_{iter}$}
      \State Extract the visual features $v_i$  by the CNN backbone ($e.g.$, ResNet101 \cite{he2016deep}) for all  samples $x_i \in X^s$;
      \State Generate virtual unseen visual feature using \eqref{eq: 2};
      \State Get the semantic embedding of real seen and virtual unseen visual features $s_i$ via the visul-semantic embedding network;
      \State Build the visual graph and semantic graph using \eqref{eq:4} and \eqref{eq:5}, respectively;
      \State Mask the built visual graph using \eqref{eq:8};
      \State Update visual graph and semantic graph using \eqref{eq:9}, \eqref{eq:10} and \eqref{eq:11};
      \State Optimize VSGMN with \eqref{eq: 30};
    \EndWhile
    \State Predict and output the labels of the samples in test set using \eqref{eq: 31}.
  \end{algorithmic}
\end{algorithm}

\subsubsection{Second-Stage Visual-Semantic Alignment} In Second-stage Visual-Semantic Alignment, the loss functions make the relationships among semantic embeddings $S_b^e$ close to the relationships among the semantic prototypes $S_b^p$. Specifically we propose the Class Relationship Constraint Loss to match categories relationships.

\noindent\textbf{Class Relationship Constraint Loss.} In GMN, we have got $L$ visual graphs and semantic graphs. Now, we need to match them to constrain the visual-semantic embedding in our GBN. Specifically, we use KL divergence to pull the probability distribution of relationships for each node in the visual graph towards its corresponding semantic graph. It is defined as:
\begin{align}
    &\mathcal{L}_{CRC}=-\frac{1}{n_b+C_u}\sum_{l=1}^{L}\sum_{i=1}^{n_b+C_u}\sum_{j=1}^{n_b+C_u}p^{sl}_{i,j} \cdot \log{(p^{sl}_{i,j}-p^{vl}_{i,j})}.
\end{align}
    Finally, we formulate the overall loss function of VSGMN:
\begin{equation}
\begin{split}
     \mathcal{L}_{total}&=\mathcal{L}_{ACE}+\lambda_{REG}\mathcal{L}_{REG} \\
                        &+\lambda_{SC}\mathcal{L}_{SC}+\lambda_{CRC}\mathcal{L}_{CRC},
\end{split}
\label{eq: 30}
\end{equation}
where $\lambda_{REG}$, $\lambda_{SC}$ and $\lambda_{CRC}$ are the weights to control their corresponding loss terms. 

\subsection{Zero-Shot Prediction}
After training VSGMN, we first obtain the embedding features of a test instance $x_i$ in the semantic space $i.e.$, ($s_i$). Then, we take an explicit calibration to predict the test label of $x_i$, which is formulated as:
\begin{equation}
    c^*=\arg \max_{c \in C^u/C} s_i \times Z^c + \mathbb{I}_{[c \in C^u]}. 
    \label{eq: 31}
\end{equation}
Here $C^u/C$ corresponds to the CZSL/GZSL setting respectively. The complete procedures (including model training and prediction) for VSGMN are illustrated by the pseudocode in Algorithm \ref{alg:1}.
\section{Experiments and Evaluation}
\label{4}
\subsection{Datasets and Evaluation Protocols}
\subsubsection{Datasets}
We evaluate our method on three challenging benchmark datasets, $i.e.$, CUB (Caltech UCSD Birds 200) \cite{welinder2010caltech}, SUN (SUN Attribute) \cite{patterson2012sun} and AWA2 (Animals with Attributes 2) \cite{xian2017zero}. Among these, CUB and SUN are fine-grained datasets, whereas AWA2 is a coarse-grained dataset. Following \cite{xian2017zero}, we use the same seen/unseen splits and class embeddings. Specifically, CUB includes 11,788 images of 200 bird classes (seen/unseen classes = 150/50) with 312 attributes. SUN has 14,340 images from 717 scene classes (seen/unseen classes = 645/72) with 102 attributes. AWA2 consists of 37,322 images from 50 animal classes (seen/unseen classes = 40/10) with 85 attributes.

\subsubsection{Evaluation Protocols} Following \cite{xian2017zero}, we evaluate the top-1 accuracy both in the CZSL and GZSL settings. In the CZSL setting, we predict the unseen classes to compute the accuracy
of test samples, $i.e.$, $acc$. In the GZSL setting, we calculate the
accuracy of the test samples from both the seen classes (denoted as
$S$) and unseen classes (denoted as $U$). Meanwhile, their harmonic
mean (defined as $H=(2 \times S \times U)/(S+U)$) is also employed for evaluating the performance in the GZSL setting.
\begin{table}[t]
\caption{Comparisons of model size and FLOPs.}
    \centering
    \setlength{\tabcolsep}{8mm}{
    \begin{tabular}{l|c|c}
         \hline
        \textbf{Method} & \textbf{Model size} & \textbf{FLOPs} \\
        \hline
        TransZero \cite{chen2022transzero}&3.31 M&22.51 G\\
        GNDAN\cite{chen24gndan}&5.77 M&12.94 G\\
        \textbf{VSGMN (prp)}&3.61 M&28.51 G\\
        \textbf{VSGMN (att)}&3.34 M&28.16 G\\
        \hline
    \end{tabular}}
    
    \label{tab: mf}
\end{table}
\subsection{Implementation Details.} We use the training splits proposed by \cite{xian2017zero}. We take a ResNet101 pre-trained on ImageNet as the CNN backbone to extract the visual feature map $v(x) \in \mathbb{R}^{H\times W \times C}$ ($H$ and $W$ are the height and width of the feature maps, $C$ is the number of channels) without fine-tuning. We use the SGD optimizer with hyperparameters (learning rate = 0.001, momentum = 0.9, weight decay = 0.0001) to optimize our model. The batch size is set to 40 for AWA2 ,and 50 for CUB and SUN. Note that we used balance sampling (only one image per class per batch) to ensure that the learned class relationships remain balanced. Following APN \cite{xu2020attribute}, hyperparameters in our model are obtained by grid search on the validation set \cite{xian2017zero}. We use PyTorch \cite{paszke2019pytorch} for the implementation of all experiments. 

Furthermore, we also supply the model size (Params) and FLOPs of different implementations and a comparison with our baseline (TransZero\cite{chen2022transzero}) and GNDAN\cite{chen24gndan} (a embedding ZSL method which also use GNNs). As shown in Table \ref{tab: mf}, due to the additional GMN module in our model, both model size and FLOPs are slightly higher compared to the baseline but remain within an acceptable range. Compared with GNDAN, our model has an advantage in model size, though the FLOPs are significantly higher. This may be due to the self-attention and cross-attention computations involved when obtaining the semantic embeddings in GBN.

\subsection{Comparison with State-of-the-Art}
Our VSGMN is embedding-based manner. We compare it with other state-of-the-art methods both in CZSL and GZSL settings including embedding-based methods ($e.g.$, SP-AEN \cite{chen2018zero}, SGMA \cite{zhu2019semantic}, AREN \cite{xie2019attentive}, LFGAA \cite{liu2019attribute}, DAZLE \cite{huynh2020fine}, APN \cite{xu2020attribute}, cvc-zsl \cite{li2019rethinking}, SR2E \cite{ge2021semantic},TransZero \cite{chen2022transzero}, MSDN \cite{chen2022msdn}, A-RSR \cite{liu24refine}, BGSNet \cite{li2023diversity}, VABNet \cite{Gao23visual}, GNDAN \cite{chen24gndan}) and Generative methods($e.g.$, SE-ZSL \cite{verma2018generalized}, f-VAEGAN \cite{xian2019f}, LisGAN \cite{li2019leveraging},AFC-GAN \cite{li2019alleviating}, OCD-CVAE \cite{keshari2020generalized}, GCM-CF \cite{yue2021counterfactual}, FREE \cite{chen2021free}, HSVA \cite{chen2021hsva}, AREES \cite{2}, D3GZSL \cite{wang2024data}), to demonstrate its effectiveness and advantages.

\begin{table*}[!t]
\caption{Results (\%) of the state-of-the-art CZSL and GZSL methods on CUB, SUN and AWA2, including generative methods and embedding-based methods. The best and second-best results are marked in Red and Blue, respectively. The symbol “–” indicates no results. The symbol “*” denotes the results are obtained from locally reproducing the method.}
    \centering
    \resizebox{\textwidth}{!}{
    \begin{tabular}{r|c|ccc|c|ccc|c|ccc}
        \hline
        \multirow{3}{*}{\textbf{Methods}} & \multicolumn{4}{c|}{\textbf{AWA2}} & \multicolumn{4}{c}{\textbf{CUB}} & \multicolumn{4}{|c}{\textbf{SUN}}    \\
        \cline{2-13}
        & CZSL & \multicolumn{3}{c|}{GZSL} & CZSL & \multicolumn{3}{c|}{GZSL} & CZSL & \multicolumn{3}{c}{GZSL} \\
        \cline{2-13}
        & acc & U&S&H&acc&U&S&H&acc&U&S&H \\
        \hline
        \textbf{Embedding-based Methods}&&&&&&&&&&&&\\
        SP-AEN\cite{chen2018zero}&58.5&23.3&90.9&37.1&55.4&34.7&70.6&46.6&59.2&24.9&38.6&30.3\\
        SGMA\cite{zhu2019semantic}&68.8&37.6&87.1&52.5&71.0&36.7&71.3&48.5&-&-&-&-\\
        AREN\cite{xie2019attentive}&67.9&15.6&\textcolor{blue}{92.9}&26.7&71.8&38.9&\textcolor{blue}{78.7}&52.1&60.6&19.0&38.8&25.5\\
        
        LFGAA\cite{liu2019attribute}&68.1&27.0&\textcolor{red}{93.4}&41.9&67.6&36.2&\textcolor{red}{80.9}&50.0&61.5&18.5&\textcolor{blue}{40.0}&25.3\\
        DAZLE\cite{huynh2020fine}&67.9&60.3&75.7&67.1&66.0&56.7&59.6&58.1&59.4&\textcolor{red}{52.3}&24.3&33.2\\
        SR2E\cite{ge2021semantic}&-&58.0&80.7&67.5&-&61.6&70.6&65.8&-&43.1&36.8&39.7\\
        APN\cite{xu2020attribute}&68.4&57.1&72.4&63.9&72.0&65.3&69.3&67.2&61.6&41.9&34.0&37.6\\
        SCILM\cite{ji2021semantic}&\textcolor{blue}{71.2}&48.9&77.8&60.1&52.3&24.5&54.9&33.8&62.4&24.8&32.6&28.2\\
        TransZero*\cite{chen2022transzero}&68.9&59.8&82.7&69.4&75.4&67.4&68.6&68.0&64.9&51.0&33.6&40.5\\
        MSDN\cite{chen2022msdn}&70.1&62.0&74.5&67.7&\textcolor{blue}{76.1}&\textcolor{blue}{68.7}&67.5&68.1&\textcolor{blue}{65.8}&\textcolor{blue}{52.2}&34.2&41.3\\
        VABNet\cite{Gao23visual}&67.6&56.1&71.8&63.0&-&-&-&-&57.0&40.1&33.4&36.4\\
        BGSNet\cite{li2023diversity}&69.1&\textcolor{blue}{61.0}&81.8&\textcolor{blue}{69.9}&73.3&60.9&73.6&66.7&63.9&45.2&34.3&39.0\\
        A-RSR\cite{liu24refine}&68.4&55.3&76.0&64.0&72.0&62.3&73.9&67.6&64.2&48.0&34.9&40.4\\
        GNDAN\cite{chen24gndan}&71.0&60.2&80.8&69.0&75.1&69.2&69.6&\textcolor{red}{69.4}&65.3&50.0&34.7&41.0\\
        \hline
        \textbf{Generative Methods}&&&&&&&&&&&&\\
        SE-ZSL\cite{verma2018generalized}&69.2&58.3&68.1&62.8&59.6&41.5&53.3&46.7&63.4&40.9&30.5&34.9\\
        f-VAEGAN\cite{xian2019f}&71.1&57.6&70.6&63.5&61.0&48.4&60.1&53.6&64.7&45.1&38.0&41.3\\
        LisGAN\cite{li2019leveraging}&70.6&52.6&76.3&62.3&58.8&46.5&57.9&51.6&61.7&42.9&37.8&40.2\\
        AFC-GAN\cite{li2019alleviating}&69.1&58.2&68.8&62.2&62.9&53.5&59.7&56.4&63.3&49.1&36.1&41.6\\
        OCD-CVAE\cite{keshari2020generalized}&-&59.5&73.4&65.7&-&44.8&59.9&51.3&-&44.8&\textcolor{red}{42.9}&\textcolor{red}{43.8}\\      
        GCM-CF\cite{yue2021counterfactual}&-&60.4&75.1&67.0&-&61.0&59.7&60.3&-&47.9&37.8&42.2\\
        FREE\cite{chen2021free}&-&60.4&75.4&67.1&-&55.7&59.9&57.7&-&47.4&37.2&41.7\\
        HSVA\cite{chen2021hsva}&-&59.3&76.6&66.8&62.8&52.7&58.3&55.3&63.8&48.6&39.0&\textcolor{blue}{43.3}\\
        AREES\cite{2}&\color{red}{73.6}&57.9&77.0&66.1&65.7&53.6&56.9&55.2&64.3&51.3&35.9&42.2\\
        D3GZSL\cite{wang2024data}&-&\textcolor{red}{64.6}&76.7&70.1&-&66.7&69.1&67.8&-&-&-&-\\
        \hline
        \textbf{VSGMN(Ours)}&\textcolor{blue}{71.2}&\textcolor{blue}{64.0}&77.8& \textcolor{red}{70.3}&\textcolor{red}{77.8}&\textcolor{red}{69.6}&68.9&\textcolor{blue}{69.3}&\textcolor{red}{66.3}&50.7&34.1&40.8\\
        \hline        
    \end{tabular}}
    
    \label{tab:1}
\end{table*}

\subsubsection{Conventional Zero-Shot Learning}
Here, we first compare our VSGMN with the state-of-the-art methods in the CZSL setting. In the three benchmark datasets, SUN and CUB are two fine-grained datasets, which means that the visual differences between categories in these two datasets are very small. However, in the semantic space, due to the discriminative nature of the attributes selected for the manually defined class prototypes, incorporating semantic categories into the visual space is crucial. As shown in Table \ref{tab:1}, our VSGMN achieves the best accuracy of 77.8\% and 66.3\% on CUB and SUN. This demonstrates that our VSGMN can introduce valuable semantic class information into visual features with very small inter-class differences in fine-grained datasets, thereby enhancing the discriminability between classes in the visual space. This also implies that our VSGMN can enhance the visual-semantic matching capability by imposing appropriate constraints on visual-semantic embedding.

As for the coarse-grained dataset AWA2, VSGMN still obtains competitive performance, with a accuracy of 71.2\%, which is the best accuracy among the embedding-based and generative-based methods. This indicates that, compared with embedding-based models that only utilize first-stage visual-semantic alignment, our VSGMN, which additionally imposes second-stage visual-semantic alignment, can more accurately learn the visual-semantic space mapping as shown in Fig \ref{fig:3}. Consequently, it optimizes the transfer of knowledge from seen classes to unseen ones. Compared with its baseline, VSGMN obtains gains of over 2.3\%, 2.4\% and 1.4\% on AWA2, CUB and SUN, respectively. This demonstrates that our proposed visual-semantic graph matching constraint method is helpful for constructing the visual space and the mapping between the visual-semantic spaces.

However, we observe that the improvement of our VSGMN on the SUN dataset (1.4\%) is less pronounced compared to AWA2 and CUB (2.3\% and 2.4\%). We hypothesize that this is because the SUN dataset contains significantly more categories than AWA2 and CUB, which leads to a more uniform probability distribution when using $\mathcal{L}_{CRC}$ for alignment. This, in turn, indirectly reduces the effectiveness of the relationship constraints.
\begin{table*}[!t]
\caption{Ablation studies for different components of VSGMN on  AWA2 and CUB datasets. baseline is the VSGMN without GMN and $\mathcal{L}_{CRC}$, “GM” denotes the graph match layer in GMN,  “UM” denotes the virtual unseen embedding mask in GMN, “CG” denotes the cross-graph message function $f_{message}^c$ in GMN}
    \centering
    \setlength{\tabcolsep}{4mm}{
    \begin{tabular}{l|c|c|c|c|c|c|c|c}
         \hline
        \multirow{3}{*}{\textbf{Method}} & \multicolumn{4}{c|}{\textbf{AWA2}} & \multicolumn{4}{c}{\textbf{CUB}} \\
        \cline{2-9}
        & CZSL & \multicolumn{3}{c|}{GZSL} & CZSL & \multicolumn{3}{c}{GZSL}  \\
        \cline{2-9}
        & acc & U&S&H&acc&U&S&H \\
        \hline
        baseline&68.9&59.8&82.7&69.4&75.4&67.4&68.6&68.0\\
        baseline + $\mathcal{L}_{CRC}$&68.6&61.2&75.8&67.7&73.3&67.3&60.9&63.9\\
        baseline + GM&70.7&63.6&77.4&69.8&76.3&67.3&69.7&68.5\\
        baseline + GM + UM&70.8&63.4&77.4&69.7&77.0&69.7&68.1&68.9\\
        baseline + GM + CG&70.9&64.2&77.3&70.1&77.1&69.3&68.3&68.8\\
        baseline + GM + CG + UM&71.2&64.0&77.8&70.3&77.8&69.6&68.9&69.3\\
        \hline
    \end{tabular}}
    
    \label{tab:2}
\end{table*}

\subsubsection{Generalized Zero-Shot Learning}
Table \ref{tab:1} also shows the results of different methods in the GZSL setting, $i.e.$, embedding-based methods and generative-based methods. We can observe that most methods tend to overfit on seen classes, resulting in the model's accuracy on recognizing seen classes much higher than that on unseen classes, which affects the performance on the H-score. However, our VSGMN can achieve a better balance between seen and unseen classes. As such, VSGMN achieves good results of Harmonic mean, $e.g.$, 70.3\% and 69.3\% on AWA2 and CUB, respectively. We believe this can be attributed to the introduction of virtual unseen class visual features and the virtual unseen embedding mask, which help GMN align the unseen relationships between visual and semantic.

Since per class only contains about 16 training images on SUN, which heavily limits the ZSL models, the data augmentation is very effective for improving the performance on SUN. Therefore, methods that generate additional samples or employ visual feature generation techniques often achieve better results on this dataset. As such, most of the strong generative methods perform
better than our VSGMN and other embedding-based methods. However, in embedding-based methods, we achieved the second-highest accuracy, second only to MSDN. We believe this is because, compared to the attention-based visual-semantic embedding network we chose, MSDN can distill semantics, allowing it to discover more intrinsic semantic representations for effective knowledge transfer from seen to unseen classes. Finally, compared with the embedding-based method and baseline, Our VSGMN has achieved significant improvements, indicating that our proposed visual-semantic graph matching model has superiority and great potential for the ZSL task.

\subsection{Ablation Study}
To provide further insight into VSGMN, we conduct ablation
studies to evaluate the effect of different model components, loss
functions,  the graph match layer implementations.

\subsubsection{Analysis of Model Components} As shown in Table \ref{tab:2}, we conduct
ablation studies to evaluate the effects of different GMN components , $i.e.$, graph match layer (denoted as GM), the cross-graph message passing function $f^c_{message}$ (denoted as CG) and virtual unseen embedding mask (denoted as UM) on AWA2 and CUB. 

Firstly, we can observe that directly adding relationship constraints to the baseline does not yield optimal results. The reason, as mentioned in Sec. \ref{3.4.1}, is that due to the inconsistency in manifold structures and the absence of higher-order relational information, directly using $\mathcal{L}_{CRC}$ to align the first-order relationships between embeddings and prototypes often fails to achieve optimal performance and may even negatively impact the model's effectiveness. Additionally, we can observe that compared to the baseline, the model using the graph match layer shows a significant improvement (1.8\%/0.4\% on AWA2 and 0.9\%/0.5 \% on CUB), indicating that our GMN, composed of multiple stacked graph match layers, successfully utilizes the visual (semantic) graph structure we built to aggregate information from both low-order and high-order neighbors onto the node representations themselves.

Furthermore, we can observe that in the version where only the virtual unseen embedding mask component was used, our VSGMN also achieved improvements in both datasets. We believe this is because the virtual unseen embedding mask can help eliminate noise from the model as mentioned above, thereby affecting performance. What's more, it can be seen that the full version of VSGMN, which incorporates both cross-graph information propagation and the virtual unseen visual feature mask, achieved the best performance on both datasets (71.2\%/70.3\% on AWA2 and 77.8\%/69.3\% on CUB), with a more significant improvement observed on the CUB dataset. We attribute this to the following two factors: i) As mentioned earlier, using only the virtual unseen embedding mask can help eliminate noise from the model, while, it also leads to a blockade in the information propagation stage. However, in the semantic graph, since the prototypes of unseen classes are real, we did not apply the mask. Therefore, the information propagation in the semantic graph is intact and smooth. By additionally applying cross-graph message passing function, we integrate the differential information of nodes into their representations. This helps alleviate the blockade while retaining the benefits of noise removal, resulting in a synergistic effect greater than the sum of its parts. ii) The CUB dataset is a fine-grained dataset where the inter-class differences in visual features are typically lower compared to the AWA2 dataset. This implies that class relationships are more critical in this context. Therefore, on the CUB dataset, the improvement brought by applying relational constraints is significantly higher compared to AWA2.


\begin{table*}[!t]
\caption{Ablation studies for different losses of VSGMN on the AWA2 and CUB datasets}
    \centering
    \setlength{\tabcolsep}{3.6mm}{
    \begin{tabular}{l|c|c|c|c|c|c|c|c}
         \hline
        \multirow{3}{*}{\textbf{Method}} & \multicolumn{4}{c|}{\textbf{AWA2}} & \multicolumn{4}{c}{\textbf{CUB}} \\
        \cline{2-9}
        & CZSL & \multicolumn{3}{c|}{GZSL} & CZSL & \multicolumn{3}{c}{GZSL}  \\
        \cline{2-9}
        & acc & U&S&H&acc&U&S&H \\
        \hline
        VSGMN w/o $\mathcal{L}_{REG}$ and $\mathcal{L}_{CRC}$&70.8&65.3&74.4&69.6&74.4&62.5&70.5&66.2\\
        VSGMN w/o $\mathcal{L}_{REG}$&70.8&65.5&74.5&69.7&75.1&62.6&70.3&66.2\\
        \hline
        VSGMN w/o $\mathcal{L}_{SC}$ and $\mathcal{L}_{CRC}$&68.2&10.5&95.7&19.0&76.3&43.0&76.8&55.1\\
        VSGMN w/o $\mathcal{L}_{SC}$&68.2&12.5&95.5&22.1&77.0&43.8&75.7&55.5\\
        \hline
        VSGMN w/o $\mathcal{L}_{CRC}$&69.1&64.4&73.2&68.5&76.0&66.3&67.7&67.0\\
        VSGMN(full)&71.2&64.0&77.8&70.3&77.8&69.6&68.9&69.3\\

        \hline
    \end{tabular}}
    
    \label{tab:3}
\end{table*}
\subsubsection{Analysis of Loss Functions}
We conducted ablation studies to evaluate the effectiveness of our second-stage visual-semantic alignment by testing different combinations of loss functions. Specifically, we tested the effectiveness of the class relationship constraint loss $\mathcal{L}_{CRC}$ in combination with different first-stage visual-semantic alignment loss functions ($\mathcal{L}_{ACE}$, $\mathcal{L}_{SC}$, $\mathcal{L}_{REG}$). Our results are shown in Table \ref{tab:3}.

Firstly, we observe that overall, regardless of how first-stage losses are combined, the additional inclusion of $\mathcal{L}_{CRC}$ yields more significant improvements on CUB compared to AWA2 (0.7\%/0.7\%/1.8\% for CZSL, 0/0.4\%/2.3\% for GZSL). This is because the CUB dataset has smaller inter-class differences and more similar class relationships. Without additional class constraints, the model itself finds it more challenging to learn optimal visual enhancement and spatial mapping. Furthermore, compared to scenarios without $\mathcal{L}_{REG}$, the inclusion of $\mathcal{L}_{CRC}$ without $\mathcal{L}_{SC}$ leads to considerable improvements in both datasets (0/3.1\%, 0.7\%/0.4\%). This is because without loss $\mathcal{L}_{CRC}$, the model is more prone to overfitting on seen classes. By adding $\mathcal{L}_{CRC}$, additional information from unseen classes is introduced, thereby enhancing the model's generalization capability to unseen classes.

Finally, we observe that under the scenario of only applying first-stage alignment losses (VSGMN w/o $\mathcal{L}_{CRC}$), the addition of class constraints leads to significant improvements on both datasets (2.1\%/1.8\% on AWA2, 1.8\%/2.3\% on CUB). This further demonstrates the effectiveness of our proposed second-stage visual-semantic alignment constraints in reducing the visual-semantic gap.

\begin{table}[]
\caption{Ablation studies for different implementations of graph match layer on the AWA and CUB datasets}
    \centering
    \setlength{\tabcolsep}{1.2mm}{
    \begin{tabular}{l|c|c|c|c|c|c|c|c}
         \hline
        \multirow{3}{*}{\textbf{Method}} & \multicolumn{4}{c|}{\textbf{AWA2}} & \multicolumn{4}{c}{\textbf{CUB}} \\
        \cline{2-9}
        & CZSL & \multicolumn{3}{c|}{GZSL} & CZSL & \multicolumn{3}{c}{GZSL}  \\
        \cline{2-9}
        & acc & U&S&H&acc&U&S&H \\
        \hline
        VSGMN (att-MLP)&70.0&63.9&73.7&68.4&75.2&69.7&62.3&65.8\\
        VSGMN (att)&71.0&59.4&86.2&70.3&77.8&69.6&68.9&69.3\\
        VSGMN (prp)&71.2&64.0&77.8&70.3&77.4&66.5&72.2&69.3\\
        \hline
    \end{tabular}}
    
    \label{tab:4}
\end{table}
\subsubsection{Analysis of Graph Match Layer Implementations} To investigate the specific architectural implementations of graph match layer and their impact on the overall performance of our VSGMN, we conducted experiments on two architectures (attention-based and propagation-based) mentioned in Section \ref{3.2.2} and different attention computation functions (MLP and dot product). The results are shown in table \ref{tab:4}. 

We can observe that the performance of the two models is very close on both the AWA2 and CUB datasets (71.0\%/71.2\%, 70.3\%/70.3\% on AWA2 and 77.8\%/77.4\%, 69.3\%/69.3\% on CUB), indicating that our second-stage visual-semantic graph alignment method is robust to the specific implementation. Additionally, we observe that in the AWA2 dataset, the propagation-based implementations slightly outperforms the attention-based method, while the opposite is observed in the CUB dataset. We attribute this to the fact that our built visual and semantic graphs do not have explicit neighbor relationships but instead utilize cosine distances to represent the proximity of relationships between nodes. The attention-based method can leverage attention scores to obtain relatively clear neighbor relationships, thus having a significant advantage in the fine-grained CUB dataset where category relationships are closer. Furthermore, we can observe that, compared to the standard attention mechanism that calculates attention scores using dot products, the MLP-based implementation is less effective. We speculate that this might be because we chose to use cosine distance to quantify category relationships in the graph matching stage. As a result, using dot products to calculate attention scores is more aligned with the optimization goal of graph matching.

\begin{figure*}[!t]
    \centering
    \includegraphics[width=\textwidth]{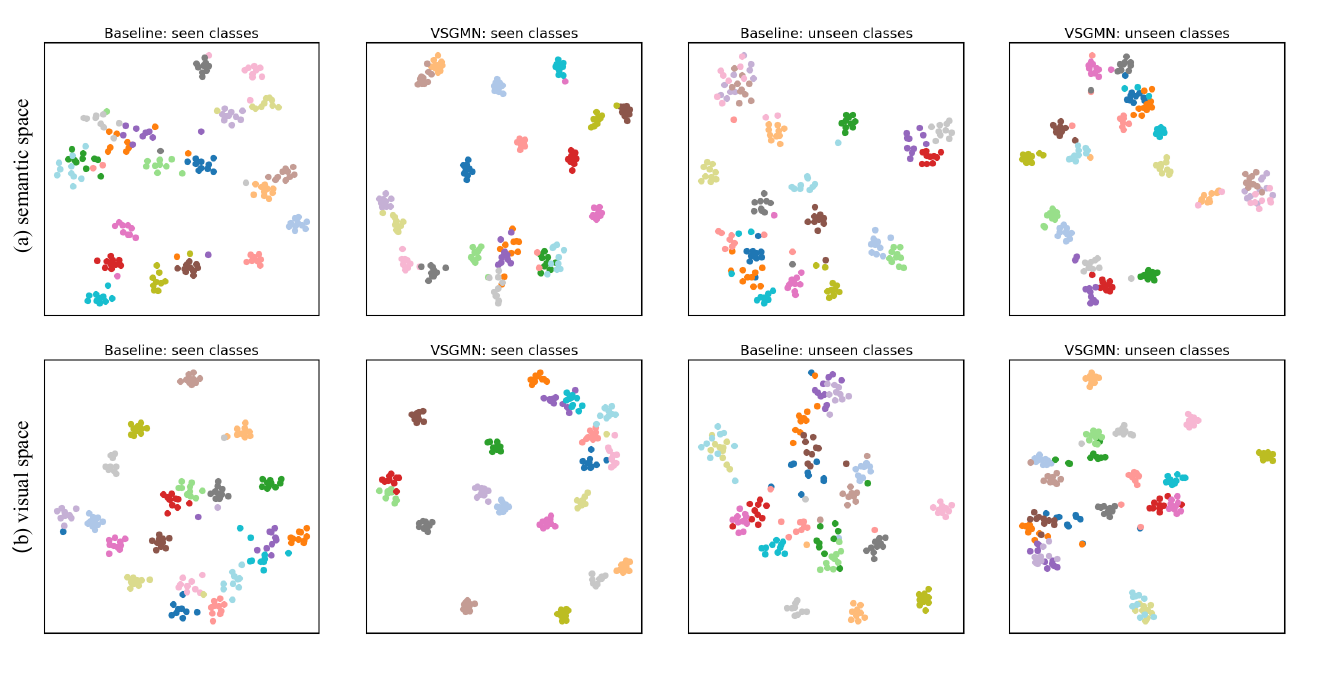}
    \caption{t-SNE visualization\cite{van2008visualizing} of visual features in semantic space and visual space learned by our VSGMN and Baseline for the same seen and unseen classes. Different colors denote different classes. We conduct experiments on 20 classes of CUB. The baseline model, while capable of finding reasonably appropriate semantic space and visual space for samples, exhibits a certain degree of category confusion, particularly evident in unseen classes. However, VSGMN notably alleviates this condition, especially concerning unseen classes.}
    \label{fig:5}
\end{figure*}
\begin{figure*}[!t]
    \centering
    \subfloat[AWA2]{ \includegraphics[width=3.5in]{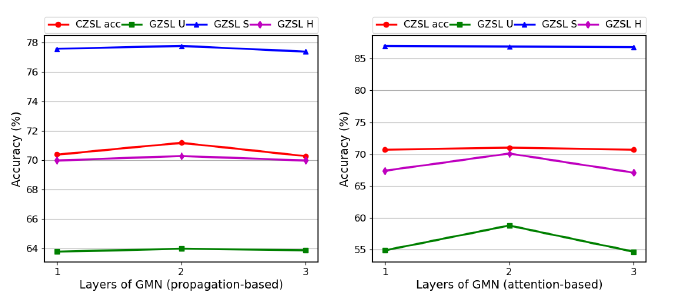} \label{fig:8.1}}
    \subfloat[CUB]{ \includegraphics[width=3.5in]{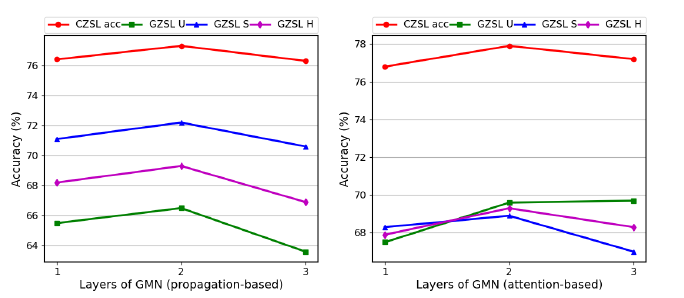} \label{fig:8.2}}
    
    \caption{: The effects of different architectures for the GMN on (a) AWA2 and (b) CUB. We investigate the number of graph match layers for propagation-based implementation and attention-based implementation.}
    \label{fig:8}
\end{figure*}
\subsection{Qualitative Results}
Here, we present the visualizations of  t-SNE \cite{van2008visualizing} in visual space and semantic space to intuitively show the effectiveness of our VSGMN.
\subsubsection{t-SNE Visualizations in Semantic Space}
As shown in Fig.\ref{fig:5} (a), we first provide the t-SNE visualization \cite{van2008visualizing} of semantic embedding for seen classes and unseen classes on CUB dataset, learned by the baseline and our VSGMN. 

Firstly, we observe that our baseline model, while capable of finding reasonably appropriate semantic space embeddings for visual features, exhibits varying degrees of category confusion on CUB. In contrast, our VSGMN, with the addition of category relation constraints, significantly mitigates this phenomenon. We attribute this to the success of our proposed second-stage visual-semantic alignment in providing appropriate class-level constraints to the space embedding network, thereby alleviating the visual-semantic gap caused by inconsistent manifold structures and enabling VSGMN to learn more accurate visual-semantic mappings. Furthermore, we found that our VSGMN can improve the intra-class similarity and inter-class distinctiveness in the semantic space for both seen and unseen classes to a certain extent. This further demonstrates that by matching visual and semantic graphs, we have successfully enhanced the overall quality of visual-semantic alignment, further confirming the effectiveness of our approach.


\subsubsection{t-SNE Visualizations in Visual Space}
In order to investigate the impact of the proposed category relation constraints on learning the visual space, we also conducted t-SNE \cite{van2008visualizing} visualization of the learned visual space for seen classes and unseen classes on CUB dataset, learned by the baseline and our VSGMN.

As shown in Fig.\ref{fig:5} (b), our baseline model exhibits category confusion in multiple classes, which can be attributed to the fact that the visual space learned by the baseline is constrained only by the first-stage visual-semantic alignment loss function mentioned in section \ref{3.5.1}. This means that our visual-semantic embedding net is limited to optimizing parameters within each class, potentially resulting in insufficient robustness of visual features due to domain shift during the lengthy optimization process. In contrast, our VSGMN introduces inter-class information to the visual semantic alignment, giving it a class-level perspective during the optimization process. This allows the model to consider the feature extraction part of all classes within a single optimization step, thereby mitigating the effects of domain shift and improving the quality of visual space learning.

\begin{figure*}[!t]
    \centering
    \includegraphics[width=\textwidth]{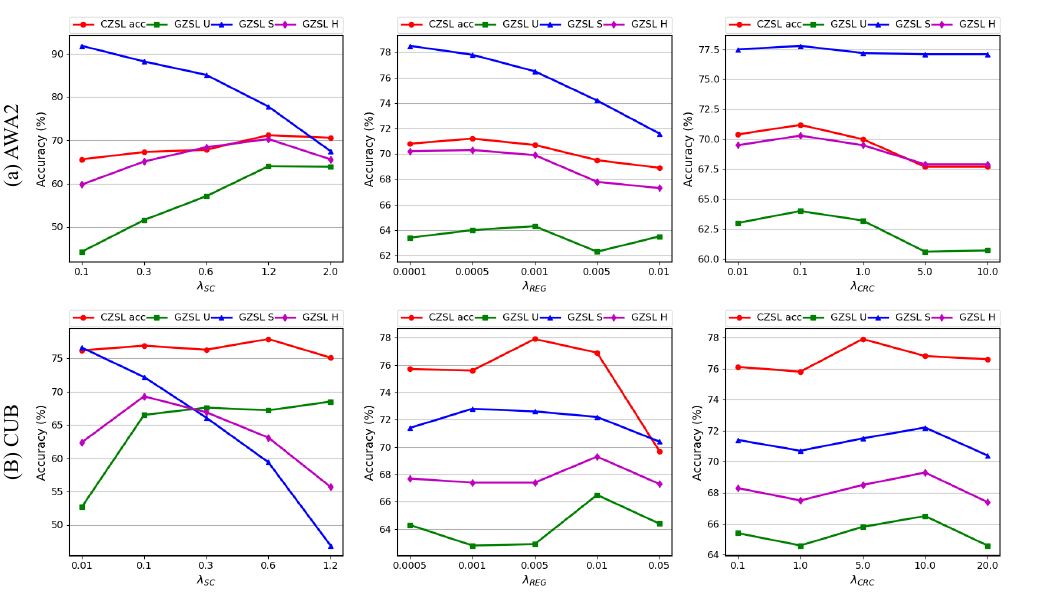}
    \caption{The effects of loss weights that control their corresponding loss terms on AWA2 and CUB, $i.e.$, $\lambda_{REG}$, $\lambda_{SC}$ and $\lambda_{CRC}$.}
    \label{fig:7}
\end{figure*}

\subsection{Hyperparameter Analysis}
\label{4.4}
To analyse the robustness of our VSGMN and select better hyperparameters for it. We conduct extensive experiments for evaluating the effects of loss weights (in \eqref{eq: 30}) and graph match layer architecture settings in GMN.

\subsubsection{Effects of Different Architectures for GMN}
To find the best graph matchine net settings, we investigate the influence of the number of layers for both graph match layer implementations. As shown in Fig.\ref{fig:8}, it can be observed that regardless of the implementation, our VSGMN achieves the best performance when the number of layers in the graph matching net is 2. This is because, compared to smaller numbers of layers ($e.g.$, 1 layer), a 2-layer GMN can aggregate higher-order neighborhood information for each node, enabling the subsequent graph matching to simultaneously align higher-order neighborhood relations, further increasing the alignment between the visual graph and the semantic graph, and reducing the visual-semantic gap. However, as the number of layers continues to increase, the performance of VSGMN decreases further. We speculate that this is due to the occurrence of over-smoothing, where node representations within the same connected component tend to converge to the same value. This phenomenon has been observed in many graph neural network-based works \cite{li2018deeper,oono2019graph,chen2020measuring}. For ZSL tasks, we hope that semantic embeddings of different categories not only have inherent category relationships but also exhibit significant differences. Therefore, the impact of over-smoothing on VSGMN is more pronounced.

\subsubsection{Effects of Loss Weights}
Here, we analyse the effects of loss weights that control their
corresponding loss terms, $i.e.$, $\lambda_{REG}$, $\lambda_{SC}$ and $\lambda_{CRC}$. We try a range of these loss weights evaluated on AWA2 and CUB. 
Whereas $\lambda_{REG}$ controls the quality of first-stage visual-semantic alignment, $\lambda_{SC}$ mitigates overfitting to seen classes effectively, and $\lambda_{CRC}$ adjusts the strength of second-stage visual-semantic alignment.

Results are shown in Fig.\ref{fig:7}. When $\lambda_{REG}$, $\lambda_{SC}$ and $\lambda_{CRC}$ are set to a large value, all evaluation protocols tend to drop. Meanwhile, VSGMN is insensitive to the self-calibration loss in the CZSL setting. Moreover, We find that on the fine-grained dataset CUB, a larger value of $\lambda_{CRC}$ is needed to achieve the best performance. This is because the class relationships in the CUB dataset are more closely related, thus requiring more assistance from class relationship constraints.

\section{Conclusion}
\label{5}
In this paper, we propose a visual-semantic graph matching network for ZSL, termed VSGMN, to achieve second-stage alignment of vision and semantics using the class relationships between semantic prototypes. Firstly, we utilize GBN to obtain visual embeddings in the semantic space, and built initial visual and semantic graphs. Considering the utilization of relationships between unseen classes, we introduce virtual visual features to the visual graph and employ virtual unseen embedding mask to minimize the flow of noise while retaining as much information about unseen relationships as possible. Then, through our proposed GMN, we incorporate information from neighboring nodes and cross-graph differences for each node in our graph, thereby integrating structural features into node representations and alleviating the visual-semantic gap caused by manifold inconsistencies. Finally, through our proposed class relationship constraint loss, we impose class-level constraints on the visual-semantic embedding to assist the overall model in achieving second-stage alignment of visual and semantic space. Extensive experiments on three popular ZSL benchmarks demonstrate the superiority of our method. 
\subsection{Limitation}
Although VSGMN has achieved promising results on three ZSL benchmark datasets, it still has some limitations. First, the GMN updates the semantic embeddings and semantic prototypes in GBN by leveraging class relationships. However, since GMN cannot access the complete visual graph during the testing phase, these updated embeddings and prototypes are not utilized in testing. Therefore, we hope to find a way to leverage class relationship information during the testing phase to assist the model in classification. Additionally, since VSGMN requires the use of additional virtual visual features, the training cost significantly increases as the number of unseen classes grows, which is another issue we need to address.

\bibliographystyle{IEEEtran} 

\begin{thebibliography}{10}
\providecommand{\url}[1]{#1}
\csname url@samestyle\endcsname
\providecommand{\newblock}{\relax}
\providecommand{\bibinfo}[2]{#2}
\providecommand{\BIBentrySTDinterwordspacing}{\spaceskip=0pt\relax}
\providecommand{\BIBentryALTinterwordstretchfactor}{4}
\providecommand{\BIBentryALTinterwordspacing}{\spaceskip=\fontdimen2\font plus
\BIBentryALTinterwordstretchfactor\fontdimen3\font minus \fontdimen4\font\relax}
\providecommand{\BIBforeignlanguage}[2]{{%
\expandafter\ifx\csname l@#1\endcsname\relax
\typeout{** WARNING: IEEEtran.bst: No hyphenation pattern has been}%
\typeout{** loaded for the language `#1'. Using the pattern for}%
\typeout{** the default language instead.}%
\else
\language=\csname l@#1\endcsname
\fi
#2}}
\providecommand{\BIBdecl}{\relax}
\BIBdecl

\bibitem{krizhevsky2012imagenet}
A.~Krizhevsky, I.~Sutskever, and G.~E. Hinton, ``Imagenet classification with deep convolutional neural networks,'' in \emph{NeurIPS}, vol.~25, 2012.

\bibitem{simonyan2014very}
K.~Simonyan and A.~Zisserman, ``Very deep convolutional networks for large-scale image recognition,'' in \emph{ICLR}, Y.~Bengio and Y.~LeCun, Eds., 2015.

\bibitem{he2016deep}
K.~He, X.~Zhang, S.~Ren, and J.~Sun, ``Deep residual learning for image recognition,'' in \emph{CVPR}, 2016, pp. 770--778.

\bibitem{vaswani2017attention}
A.~Vaswani, N.~Shazeer, N.~Parmar, J.~Uszkoreit, L.~Jones, A.~N. Gomez, {\L}.~Kaiser, and I.~Polosukhin, ``Attention is all you need,'' in \emph{NeurIPS}, vol.~30, 2017.

\bibitem{larochelle2008zero}
H.~Larochelle, D.~Erhan, and Y.~Bengio, ``Zero-data learning of new tasks.'' in \emph{AAAI}, vol.~1, no.~2, 2008, p.~3.

\bibitem{palatucci2009zero}
M.~Palatucci, D.~Pomerleau, G.~E. Hinton, and T.~M. Mitchell, ``Zero-shot learning with semantic output codes,'' in \emph{NeurIPS}, vol.~22, 2009.

\bibitem{lampert2009learning}
C.~H. Lampert, H.~Nickisch, and S.~Harmeling, ``Learning to detect unseen object classes by between-class attribute transfer,'' in \emph{CVPR}, 2009, pp. 951--958.

\bibitem{lampert2013attribute}
C.~H. \vspace{0mm}Lampert, H.~Nickisch, and S.~Harmeling, ``Attribute-based classification for zero-shot visual object categorization,'' \emph{IEEE transactions on pattern analysis and machine intelligence}, vol.~36, no.~3, pp. 453--465, 2013.

\bibitem{fu2015transductive}
Y.~Fu, T.~M. Hospedales, T.~Xiang, and S.~Gong, ``Transductive multi-view zero-shot learning,'' \emph{IEEE transactions on pattern analysis and machine intelligence}, vol.~37, no.~11, pp. 2332--2345, 2015.

\bibitem{fu2017zero}
Z.~Fu, T.~Xiang, E.~Kodirov, and S.~Gong, ``Zero-shot learning on semantic class prototype graph,'' \emph{IEEE transactions on pattern analysis and machine intelligence}, vol.~40, no.~8, pp. 2009--2022, 2017.

\bibitem{socher2013zero}
R.~Socher, M.~Ganjoo, C.~D. Manning, and A.~Ng, ``Zero-shot learning through cross-modal transfer,'' in \emph{NeurIPS}, vol.~26, 2013.

\bibitem{xian2017zero}
Y.~Xian, B.~Schiele, and Z.~Akata, ``Zero-shot learning-the good, the bad and the ugly,'' in \emph{CVPR}, 2017, pp. 4582--4591.

\bibitem{schonfeld2019generalized}
E.~Schonfeld, S.~Ebrahimi, S.~Sinha, T.~Darrell, and Z.~Akata, ``Generalized zero-and few-shot learning via aligned variational autoencoders,'' in \emph{CVPR}, 2019, pp. 8247--8255.

\bibitem{chen2021hsva}
S.~Chen, G.~Xie, Y.~Liu, Q.~Peng, B.~Sun, H.~Li, X.~You, and L.~Shao, ``Hsva: Hierarchical semantic-visual adaptation for zero-shot learning,'' in \emph{NeurIPS}, vol.~34, 2021, pp. 16\,622--16\,634.

\bibitem{xian2019f}
Y.~Xian, S.~Sharma, B.~Schiele, and Z.~Akata, ``f-vaegan-d2: A feature generating framework for any-shot learning,'' in \emph{CVPR}, 2019, pp. 10\,275--10\,284.

\bibitem{yu2020episode}
Y.~Yu, Z.~Ji, J.~Han, and Z.~Zhang, ``Episode-based prototype generating network for zero-shot learning,'' in \emph{CVPR}, 2020, pp. 14\,035--14\,044.

\bibitem{chen2021free}
S.~Chen, W.~Wang, B.~Xia, Q.~Peng, X.~You, F.~Zheng, and L.~Shao, ``Free: Feature refinement for generalized zero-shot learning,'' in \emph{ICCV}, 2021, pp. 122--131.

\bibitem{chen2023egans}
S.~Chen, S.~Chen, W.~Hou, W.~Ding, and X.~You, ``Egans: Evolutionary generative adversarial network search for zero-shot learning,'' \emph{IEEE Transactions on Evolutionary Computation}, 2023.

\bibitem{hong2022semantic}
Z.~Hong, S.~Chen, G.-S. Xie, W.~Yang, J.~Zhao, Y.~Shao, Q.~Peng, and X.~You, ``Semantic compression embedding for generative zero-shot learning.'' in \emph{IJCAI}, 2022, pp. 956--963.

\bibitem{chen2023evolving}
S.~Chen, W.~Hou, Z.~Hong, X.~Ding, Y.~Song, X.~You, T.~Liu, and K.~Zhang, ``Evolving semantic prototype improves generative zero-shot learning,'' in \emph{ICML}, 2023, pp. 4611--4622.

\bibitem{hou2024visual}
W.~Hou, S.~Chen, S.~Chen, Z.~Hong, Y.~Wang, X.~Feng, S.~Khan, F.~S. Khan, and X.~You, ``Visual-augmented dynamic semantic prototype for generative zero-shot learning,'' in \emph{CVPR}, 2024, pp. 23\,627--23\,637.

\bibitem{chen2022msdn}
S.~Chen, Z.~Hong, G.-S. Xie, W.~Yang, Q.~Peng, K.~Wang, J.~Zhao, and X.~You, ``Msdn: Mutually semantic distillation network for zero-shot learning,'' in \emph{CVPR}, 2022, pp. 7612--7621.

\bibitem{chen2022transzero}
S.~Chen, Z.~Hong, Y.~Liu, G.-S. Xie, B.~Sun, H.~Li, Q.~Peng, K.~Lu, and X.~You, ``Transzero: Attribute-guided transformer for zero-shot learning,'' in \emph{AAAI}, vol.~36, no.~1, 2022, pp. 330--338.

\bibitem{xie2020region}
G.-S. Xie, L.~Liu, F.~Zhu, F.~Zhao, Z.~Zhang, Y.~Yao, J.~Qin, and L.~Shao, ``Region graph embedding network for zero-shot learning,'' in \emph{ECCV}, 2020, pp. 562--580.

\bibitem{wang2018zero}
X.~Wang, Y.~Ye, and A.~Gupta, ``Zero-shot recognition via semantic embeddings and knowledge graphs,'' in \emph{CVPR}, 2018, pp. 6857--6866.

\bibitem{li2019rethinking}
K.~Li, M.~R. Min, and Y.~Fu, ``Rethinking zero-shot learning: A conditional visual classification perspective,'' in \emph{ICCV}, 2019, pp. 3583--3592.

\bibitem{welinder2010caltech}
P.~Welinder, S.~Branson, T.~Mita, C.~Wah, F.~Schroff, S.~Belongie, and P.~Perona, ``Caltech-ucsd birds 200,'' 2010.

\bibitem{patterson2012sun}
G.~Patterson and J.~Hays, ``Sun attribute database: Discovering, annotating, and recognizing scene attributes,'' in \emph{CVPR}, 2012, pp. 2751--2758.

\bibitem{song2018transductive}
J.~Song, C.~Shen, Y.~Yang, Y.~Liu, and M.~Song, ``Transductive unbiased embedding for zero-shot learning,'' in \emph{CVPR}, 2018, pp. 1024--1033.

\bibitem{li2018discriminative}
Y.~Li, J.~Zhang, J.~Zhang, and K.~Huang, ``Discriminative learning of latent features for zero-shot recognition,'' in \emph{CVPR}, 2018, pp. 7463--7471.

\bibitem{xian2018feature}
Y.~Xian, T.~Lorenz, B.~Schiele, and Z.~Akata, ``Feature generating networks for zero-shot learning,'' in \emph{CVPR}, 2018, pp. 5542--5551.

\bibitem{min2020domain}
S.~Min, H.~Yao, H.~Xie, C.~Wang, Z.-J. Zha, and Y.~Zhang, ``Domain-aware visual bias eliminating for generalized zero-shot learning,'' in \emph{CVPR}, 2020, pp. 12\,664--12\,673.

\bibitem{xie2019attentive}
G.-S. Xie, L.~Liu, X.~Jin, F.~Zhu, Z.~Zhang, J.~Qin, Y.~Yao, and L.~Shao, ``Attentive region embedding network for zero-shot learning,'' in \emph{CVPR}, 2019, pp. 9384--9393.

\bibitem{huynh2020fine}
D.~Huynh and E.~Elhamifar, ``Fine-grained generalized zero-shot learning via dense attribute-based attention,'' in \emph{CVPR}, 2020, pp. 4483--4493.

\bibitem{chen24gndan}
S.~Chen, Z.~Hong, G.~Xie, Q.~Peng, X.~You, W.~Ding, and L.~Shao, ``Gndan: Graph navigated dual attention network for zero-shot learning,'' \emph{IEEE Transactions on Neural Networks and Learning Systems}, vol.~35, no.~4, pp. 4516--4529, 2024.

\bibitem{chen2022transzero++}
S.~Chen, Z.~Hong, W.~Hou, G.-S. Xie, Y.~Song, J.~Zhao, X.~You, S.~Yan, and L.~Shao, ``Transzero++: Cross attribute-guided transformer for zero-shot learning,'' \emph{IEEE transactions on pattern analysis and machine intelligence}, 2022.

\bibitem{ge2022dual}
J.~Ge, H.~Xie, S.~Min, P.~Li, and Y.~Zhang, ``Dual part discovery network for zero-shot learning,'' in \emph{ACM MM}, 2022, pp. 3244--3252.

\bibitem{chen2024rethinking}
S.~Chen, S.~Chen, G.-S. Xie, X.~Shu, X.~You, and X.~Li, ``Rethinking attribute localization for zero-shot learning,'' \emph{Science China Information Sciences}, vol.~67, no.~7, p. 172103, 2024.

\bibitem{chen2024progressive}
S.~Chen, W.~Hou, S.~Khan, and F.~S. Khan, ``Progressive semantic-guided vision transformer for zero-shot learning,'' in \emph{CVPR}, 2024, pp. 23\,964--23\,974.

\bibitem{chen2024causal}
S.~Chen, D.~Fu, S.~Chen, W.~Hou, X.~You \emph{et~al.}, ``Causal visual-semantic correlation for zero-shot learning,'' in \emph{ACM MM}, 2024.

\bibitem{guo2023graph}
J.~Guo, S.~Guo, Q.~Zhou, Z.~Liu, X.~Lu, and F.~Huo, ``Graph knows unknowns: Reformulate zero-shot learning as sample-level graph recognition,'' in \emph{AAAI}, vol.~37, no.~6, 2023, pp. 7775--7783.

\bibitem{frome2013devise}
A.~Frome, G.~S. Corrado, J.~Shlens, S.~Bengio, J.~Dean, M.~Ranzato, and T.~Mikolov, ``Devise: A deep visual-semantic embedding model,'' in \emph{NeurIPS}, vol.~26, 2013.

\bibitem{wang2017zero}
Q.~Wang and K.~Chen, ``Zero-shot visual recognition via bidirectional latent embedding,'' \emph{International Journal of Computer Vision}, vol. 124, pp. 356--383, 2017.

\bibitem{hubert2017learning}
Y.-H. Hubert~Tsai, L.-K. Huang, and R.~Salakhutdinov, ``Learning robust visual-semantic embeddings,'' in \emph{ICCV}, 2017, pp. 3571--3580.

\bibitem{liu2018generalized}
S.~Liu, M.~Long, J.~Wang, and M.~I. Jordan, ``Generalized zero-shot learning with deep calibration network,'' in \emph{NeurIPS}, vol.~31, 2018.

\bibitem{pourpanah2022review}
F.~Pourpanah, M.~Abdar, Y.~Luo, X.~Zhou, R.~Wang, C.~P. Lim, X.-Z. Wang, and Q.~J. Wu, ``A review of generalized zero-shot learning methods,'' \emph{IEEE transactions on pattern analysis and machine intelligence}, vol.~45, no.~4, pp. 4051--4070, 2022.

\bibitem{gori2005new}
M.~Gori, G.~Monfardini, and F.~Scarselli, ``A new model for learning in graph domains,'' in \emph{IJCNN}, vol.~2, 2005, pp. 729--734.

\bibitem{hamilton2017inductive}
W.~Hamilton, Z.~Ying, and J.~Leskovec, ``Inductive representation learning on large graphs,'' in \emph{NeurIPS}, vol.~30, 2017.

\bibitem{bruna2013spectral}
J.~Bruna, W.~Zaremba, A.~Szlam, and Y.~LeCun, ``Spectral networks and locally connected networks on graphs,'' in \emph{ICLR}, Y.~Bengio and Y.~LeCun, Eds., 2014.

\bibitem{defferrard2016convolutional}
M.~Defferrard, X.~Bresson, and P.~Vandergheynst, ``Convolutional neural networks on graphs with fast localized spectral filtering,'' in \emph{NeurIPS}, vol.~29, 2016.

\bibitem{velickovic2017graph}
P.~Velickovic, G.~Cucurull, A.~Casanova, A.~Romero, P.~Lio, Y.~Bengio \emph{et~al.}, ``Graph attention networks,'' \emph{stat}, vol. 1050, no.~20, pp. 10--48\,550, 2017.

\bibitem{pennington2014glove}
J.~Pennington, R.~Socher, and C.~D. Manning, ``Glove: Global vectors for word representation,'' in \emph{EMNLP}, 2014, pp. 1532--1543.

\bibitem{santoro2016meta}
A.~Santoro, S.~Bartunov, M.~Botvinick, D.~Wierstra, and T.~Lillicrap, ``Meta-learning with memory-augmented neural networks,'' in \emph{ICML}, 2016, pp. 1842--1850.

\bibitem{snell2017prototypical}
J.~Snell, K.~Swersky, and R.~Zemel, ``Prototypical networks for few-shot learning,'' in \emph{NeurIPS}, vol.~30, 2017.

\bibitem{liu2021isometric}
L.~Liu, T.~Zhou, G.~Long, J.~Jiang, X.~Dong, and C.~Zhang, ``Isometric propagation network for generalized zero-shot learning,'' in \emph{ICLR}, 2021.

\bibitem{zhang2023attribute}
H.~Zhang, M.~Liu, Y.~Li, M.~Yan, Z.~Gao, X.~Chang, and L.~Nie, ``Attribute-guided collaborative learning for partial person re-identification,'' \emph{IEEE Transactions on Pattern Analysis and Machine Intelligence}, 2023.

\bibitem{li2019graph}
Y.~Li, C.~Gu, T.~Dullien, O.~Vinyals, and P.~Kohli, ``Graph matching networks for learning the similarity of graph structured objects,'' in \emph{ICML}, 2019, pp. 3835--3845.

\bibitem{zhu2019semantic}
Y.~Zhu, J.~Xie, Z.~Tang, X.~Peng, and A.~Elgammal, ``Semantic-guided multi-attention localization for zero-shot learning,'' in \emph{NeurIPS}, vol.~32, 2019.

\bibitem{xu2020attribute}
W.~Xu, Y.~Xian, J.~Wang, B.~Schiele, and Z.~Akata, ``Attribute prototype network for zero-shot learning,'' in \emph{NeurIPS}, vol.~33, 2020, pp. 21\,969--21\,980.

\bibitem{paszke2019pytorch}
A.~Paszke, S.~Gross, F.~Massa, A.~Lerer, J.~Bradbury, G.~Chanan, T.~Killeen, Z.~Lin, N.~Gimelshein, L.~Antiga \emph{et~al.}, ``Pytorch: An imperative style, high-performance deep learning library,'' in \emph{NeurIPS}, vol.~32, 2019.

\bibitem{chen2018zero}
L.~Chen, H.~Zhang, J.~Xiao, W.~Liu, and S.-F. Chang, ``Zero-shot visual recognition using semantics-preserving adversarial embedding networks,'' in \emph{CVPR}, 2018, pp. 1043--1052.

\bibitem{liu2019attribute}
Y.~Liu, J.~Guo, D.~Cai, and X.~He, ``Attribute attention for semantic disambiguation in zero-shot learning,'' in \emph{ICCV}, 2019, pp. 6698--6707.

\bibitem{ge2021semantic}
J.~Ge, H.~Xie, S.~Min, and Y.~Zhang, ``Semantic-guided reinforced region embedding for generalized zero-shot learning,'' in \emph{AAAI}, vol.~35, no.~2, 2021, pp. 1406--1414.

\bibitem{liu24refine}
Z.~Liu, Y.~Li, L.~Yao, J.~McAuley, and S.~Dixon, ``Rethink, revisit, revise: A spiral reinforced self-revised network for zero-shot learning,'' \emph{IEEE Transactions on Neural Networks and Learning Systems}, vol.~35, no.~1, pp. 657--669, 2024.

\bibitem{li2023diversity}
Y.~Li, Z.~Liu, X.~Chang, J.~McAuley, and L.~Yao, ``Diversity-boosted generalization-specialization balancing for zero-shot learning,'' \emph{IEEE Transactions on Multimedia}, 2023.

\bibitem{Gao23visual}
R.~Gao, X.~Hou, J.~Qin, Y.~Shen, Y.~Long, L.~Liu, Z.~Zhang, and L.~Shao, ``Visual-semantic aligned bidirectional network for zero-shot learning,'' \emph{IEEE Transactions on Multimedia}, vol.~25, pp. 1649--1664, 2023.

\bibitem{verma2018generalized}
V.~K. Verma, G.~Arora, A.~Mishra, and P.~Rai, ``Generalized zero-shot learning via synthesized examples,'' in \emph{CVPR}, 2018, pp. 4281--4289.

\bibitem{li2019leveraging}
J.~Li, M.~Jing, K.~Lu, Z.~Ding, L.~Zhu, and Z.~Huang, ``Leveraging the invariant side of generative zero-shot learning,'' in \emph{CVPR}, 2019, pp. 7402--7411.

\bibitem{li2019alleviating}
J.~Li, M.~Jing, K.~Lu, L.~Zhu, Y.~Yang, and Z.~Huang, ``Alleviating feature confusion for generative zero-shot learning,'' in \emph{ACM MM}, 2019, pp. 1587--1595.

\bibitem{keshari2020generalized}
R.~Keshari, R.~Singh, and M.~Vatsa, ``Generalized zero-shot learning via over-complete distribution,'' in \emph{CVPR}, 2020, pp. 13\,300--13\,308.

\bibitem{yue2021counterfactual}
Z.~Yue, T.~Wang, Q.~Sun, X.-S. Hua, and H.~Zhang, ``Counterfactual zero-shot and open-set visual recognition,'' in \emph{CVPR}, 2021, pp. 15\,404--15\,414.

\bibitem{2}
Y.~Liu, Y.~Dang, X.~Gao, J.~Han, and L.~Shao, ``Zero-shot learning with attentive region embedding and enhanced semantics,'' \emph{IEEE Transactions on Neural Networks and Learning Systems}, vol.~35, no.~3, pp. 4220--4231, 2024.

\bibitem{wang2024data}
Y.~Wang, M.~Hong, L.~Huangfu, and S.~Huang, ``Data distribution distilled generative model for generalized zero-shot recognition,'' in \emph{AAAI}, vol.~38, no.~6, 2024, pp. 5695--5703.

\bibitem{ji2021semantic}
Z.~Ji, X.~Yu, Y.~Yu, Y.~Pang, and Z.~Zhang, ``Semantic-guided class-imbalance learning model for zero-shot image classification,'' \emph{IEEE Transactions on Cybernetics}, vol.~52, no.~7, pp. 6543--6554, 2021.

\bibitem{van2008visualizing}
L.~Van~der Maaten and G.~Hinton, ``Visualizing data using t-sne.'' \emph{Journal of machine learning research}, vol.~9, no.~11, 2008.

\bibitem{li2018deeper}
Q.~Li, Z.~Han, and X.-M. Wu, ``Deeper insights into graph convolutional networks for semi-supervised learning,'' in \emph{AAAI}, vol.~32, no.~1, 2018.

\bibitem{oono2019graph}
K.~Oono and T.~Suzuki, ``Graph neural networks exponentially lose expressive power for node classification,'' in \emph{ICLR}, 2020.

\bibitem{chen2020measuring}
D.~Chen, Y.~Lin, W.~Li, P.~Li, J.~Zhou, and X.~Sun, ``Measuring and relieving the over-smoothing problem for graph neural networks from the topological view,'' in \emph{AAAI}, vol.~34, no.~04, 2020, pp. 3438--3445.

\end{thebibliography}


\end{document}
\endinput